\documentclass[10pt, a4paper]{article}
\usepackage{lrec2022} 
\usepackage{multibib}
\newcites{languageresource}{Language Resources}
\usepackage{graphicx}
\usepackage{tabularx}
\usepackage{soul}
\usepackage{titlesec}
\titleformat{\section}{\normalfont\large\bfseries\center}{\thesection.}{1em}{}
\titleformat{\subsection}{\normalfont\SmallTitleFont\bfseries\raggedright}{\thesubsection.}{1em}{}
\titleformat{\subsubsection}{\normalfont\normalsize\bfseries\raggedright}{\thesubsubsection.}{1em}{}
\renewcommand\thesection{\arabic{section}}
\renewcommand\thesubsection{\thesection.\arabic{subsection}}
\renewcommand\thesubsubsection{\thesubsection.\arabic{subsubsection}}

\usepackage{epstopdf}
\usepackage[utf8]{inputenc}

\usepackage{hyperref}
\usepackage{xstring}

\usepackage{color}
\usepackage{xcolor}

\usepackage{booktabs}

\newcommand{\de}[0]{DE}
\newcommand{\fr}[0]{FR}
\newcommand{\mphasis}[0]{M-Phasis}
\newcommand{\translation}[1]{\textcolor{gray}{\small{(#1)}}}
\newcommand{\textttm}[1]{${\tt #1}$}
\newcommand\blfootnote[1]{%
  \begingroup
  \renewcommand\thefootnote{}\footnote{#1}%
  \addtocounter{footnote}{-1}%
  \endgroup
}

\title{Placing \mphasis\ on the Plurality of Hate: \\
A Feature-Based Corpus of Hate Online}

\name{Dana Ruiter$^{\dagger\ast}$, Liane Reiners$^{\ddagger\ast}$, Ashwin Geet D'Sa$^{\clubsuit}$, Thomas Kleinbauer$^{\dagger}$,\\
{\bf \large Dominique Fohr$^{\clubsuit}$, Irina Illina$^{\clubsuit}$, Dietrich Klakow$^{\dagger}$, Christian Schemer$^{\ddagger}$, Angeliki Monnier$^{\spadesuit}$}}

\address{$^{\dagger}$Spoken Language Systems Group, Saarland University, Germany \\
        \{druiter, kleiba, dietrich.klakow\}@lsv.uni-saarland.de \\ \\
        $^{\ddagger}$Department of Communication, Johannes Gutenberg University Mainz \\
        \{liane.reiners, schemer\}@uni-mainz.de \\ \\
        $^{\clubsuit}$Université de Lorraine, CNRS, Inria, LORIA \\
        \{ashwin-geet.dsa, dominique.fohr, irina.illina\}@loria.fr \\ \\
        $^{\spadesuit}$ Université de Lorraine, CREM \\
         angeliki.monnier@univ-lorraine.fr\\}

\abstract{
Even though hate speech (HS) online has been an important object of research in the last decade, most HS-related corpora over-simplify the phenomenon of hate by attempting to label user comments as \emph{hate} or \emph{neutral}. This ignores the complex and subjective nature of HS, which limits the real-life applicability of classifiers trained on these corpora.
In this study, we present the \mphasis\ corpus, a corpus of $\sim9k$ German and French user comments collected from migration-related news articles. It goes beyond the \textit{hate}-\textit{neutral} dichotomy and is instead annotated with $23$ features, which in combination become descriptors of various types of speech, ranging from critical comments to implicit and explicit expressions of hate. The annotations are performed by 4 native speakers per language and achieve high ($0.77 \leq \kappa \leq 1$) inter-annotator agreements.
Besides describing the corpus creation and presenting insights from a content, error and domain analysis, we explore its data characteristics by training several classification baselines. 
 \\ \newline  }

\begin{document}

\maketitleabstract

\section{Introduction}

The internet has made the exchange of information and ideas between individuals easier than ever before. But through the provision of anonymity and filter bubbles, it has also contributed to the propagation of hateful contents that are a threat to the open exchange of opinions.
To gain insights into the dynamics and characteristics of different types of hateful speech and to develop counter-measures, communication researchers and computer scientists alike require high-quality annotated data.\blfootnote{$^{\ast}$ Equal contribution.}

To date, published datasets related to hate speech (HS) come with several limitations. 
Most corpora used to train HS classifiers over-simplify the phenomenon of HS by labelling user content with binary classes, e.g., \emph{hate}/\emph{neutral}, whose underlying definition varies greatly across corpora \cite{jurgens-etal-2019-just}. Another common limitation in HS corpora is the use of slur \cite{rost2016digital} or emotion word lists \cite{paltoglou2013predicting} to identify and sample hateful content. These approaches are insensitive to more subtle forms of HS (i.e., implicit hate) that is conveyed through syntactical or contextual features \cite{Cohen-Almagor_2018}.
Further, most hate speech corpora are based on the same outlet, i.e., Twitter (more than $50\%$ of datasets), and language, i.e., English ($\sim40\%$) \cite{vidgen2021directions}. 
Focusing on Twitter data is especially limiting, given that Twitter poses a special case of user interactions due to the small maximum length of tweets. The English-centrism ignores the cross-cultural differences of the manifestations of hate.
Also, HS corpora often do not provide access to the conversational context in which a comment is embedded, which is problematic since HS highly depends on the context \cite{kovacs2021challenges}. 
All of these limitations reduce the generalisability of HS analysis and classifiers.

In this paper, we present \textit{\textbf{M}igration and \textbf{P}atterns of \textbf{Ha}te \textbf{S}peech \textbf{i}n \textbf{S}ocial Media} (\mphasis), a corpus which focuses on the topic of migration and addresses several of the above limitations of current HS corpora. Our contributions are the following:

\begin{enumerate}
    \item Collection (Section \ref{s:corpus}) of user comments which: $1)$ are sampled from news articles that match \textbf{migration-related} regular expression keywords to ensure relevance to the topic of migration; $2)$ are \underline{not} sampled based on a list of pre-defined slur keywords and thus also capture \textbf{implicit} forms of hate; $3)$ are derived from a \textbf{diverse} set of mainstream and fringe media outlets; $4)$ are in two languages, \textbf{French and German}, for which only few HS corpora exist; $5)$ are collected as a comment thread to allow \textbf{context-sensitive} analysis.
    \item Comment annotations are based on \textbf{features} such as negative and positive evaluations,
    contrasting of groups, and expressions of emotion, which in combination become descriptors of various types of speech, ranging from critical comments to implicit and explicit hate.
    \item To identify difficulties of the \mphasis\ corpus and to provide a guide for future research, we train and evaluate \textbf{classification baselines} on it (Section \ref{s:baselines}) and analyse errors (Section \ref{s:qualitative_error_analysis}).
    \item \textbf{Analysis} of $1)$ the frequent agent-victim tuples found in the corpus (Section \ref{s:tuples}) as well as $2)$ of the domain differences between comments of different media outlets (Section \ref{s:domain_analysis}).
\end{enumerate}

\section{Related Work}

HS classifiers that detect abusive content online and flag it for human moderation or automatic deletion are the most common \textbf{computational approach} to counter HS online \cite{jurgens-etal-2019-just}. These classifiers are furthermore important research tools, e.g., to explore the dynamics of specific types of HS online \cite{johnson2019hidden,uyheng2021characterizing} or to identify common targets of abuse that require special protection \cite{silva2021analyzing}. The algorithms behind HS classifiers are manifold \cite{schmidt-wiegand-2017-survey}, ranging from statistical machine learning methods \cite{saleem_web_2016,waseem2016hateful} to neural approaches applying 
representations of language models \cite{yang2019xlnet} in single or multi-task \cite{plazadelarco2021multitask} settings.

In \textbf{social sciences}, the focus of HS research lies on the analysis of the manifestation of hate, its dynamics and role in society. A common approach is quantitative content analysis. It focuses on the investigation of manifest media content in a systematic, objective and quantitative fashion \cite{berelson1952content}. Therefore, an extensive annotation protocol is developed.
These annotations are more extensive than those typically performed in computer science, and often also take into account the context.
Social science distinguishes between different forms of impolite, uncivil or intolerant communication \cite{coe2014online,su2018uncivil,rossini2020beyond}; more fine-grained than the binary distinction commonly used in HS corpora. What distinguishes HS particularly from other concepts is that the hateful expression is group-oriented \cite{erjavec2012you}. 
Often content analyses treat HS as a special form of incivility \cite{ziegele2018socially} or harmful speech \cite{faris2016understanding} without investigating it further. But there exist also exclusive HS content analyses focusing on e.g., racist speech \cite{harlow2015story}, gendered HS \cite{doring2020gendered} or HS targeting refugees and immigrants \cite{paasch2021insult}.

A \textbf{hate speech corpus} that satisfies the different needs of computer and communication scientists requires quantity (to be able to learn detection) and granularity (to analyse various facets of HS).
Due to the different research questions addressed in communication science, the granularity-focused corpora are usually not published. This reduces the reproducibility and constantly forces researchers to create their own data annotations, which is money and time consuming.
The vast majority of published HS corpora thus favour quantity over granularity, which come with various known limitations.
Firstly, most HS corpora focus on a binary classification, e.g., \emph{hate} or \emph{non-hate} \cite{azalden2018dataset}, whose underlying meaning varies across corpora based on their annotation protocols. Depending on the focus of the HS corpus, the annotated classes vary greatly \cite{vidgen2021directions}, ranging from: person-directed abuse (e.g., cyber bullying) \cite{wulczyn2017exmachina,sprugnoli-etal-2018-creating} to group-directed abuse such as sexism \cite{jha-mamidi-2017-compliment} or racism \cite{waseem2016hateful,gudbjartur2019offensive}.
This diversity of class definitions makes it difficult to effectively combine corpora to train classifiers that generalise well across similar HS tasks \cite{ruiter2019lsv,bose-etal-2021-unsupervised}.
Further, the binarisation (e.g., \emph{sexist/not-sexist}) of HS phenomena often leads to classifiers that are unreliable and/or biased \cite{wiegand-etal-2019-detection}. More recent corpora try to overcome this limitation by creating tasks of higher granularity, focusing on multi-class tasks which may describe the target type (group vs. individual) or intensity of the abuse \cite{ousidhoum-etal-2019-multilingual}. 
\newcite{basile-etal-2019-semeval} also annotate the aggressiveness of the abuse, focusing on migrants and women. The multi-class approach with a focus on migration makes this corpus the closest to our work.
Overall there is a trend towards more complex annotations, but most approaches (including \newcite{basile-etal-2019-semeval}) still attempt to make judgements about what constitutes hate, which stands in contrast to the complex and subjective nature of HS.

We overcome the difficulty of objectively defining HS by moving beyond judgements of whether a statement is hateful or not. With the content-analytical approach in mind, we focus on annotating HS-related features, a procedure  similar to the one of \newcite{paasch2021insult}. 
Further, the \mphasis\ corpus is based on user content posted on a variety of mainstream and fringe media platforms in two languages:
French (\fr) \cite{chung-etal-2019-conan,ousidhoum-etal-2019-multilingual} and German (\de)
\cite{Bretschneider2017Detecting,struss2019overview,majumder2019overview}.

\section{The \mphasis\ Corpus}
\label{s:corpus}

\subsection{Dataset Collection}

\paragraph{Choice of Outlets} 
Instead of focusing on a single data source such as Twitter, we keep our sources diverse by focusing on user content posted in comment sections of several popular news outlets in France and Germany. To cover a broad political spectrum, we focus on four mainstream news outlets and two/three fringe media outlets in France and Germany, respectively. Concretely, the French data is collected from mainstream outlets \textit{France Info} (\emph{fi}), \textit{Le Figaro} (\emph{lf}), \textit{Le Monde} (\emph{lm}), \textit{Valeurs Actuelles} (\emph{va}) and fringe media \textit{AgoraVox} (\emph{av}), \textit{Riposte Laïque} (\emph{rl}), while the German data stems from mainstream \textit{Tagesschau} (\emph{ts}), \textit{Welt} (\emph{we}), \textit{Zeit} (\emph{ze}), \textit{Focus} (\emph{fo}) and fringe \textit{Compact} (\emph{co}), \textit{Epoch Times} (\emph{et}), \textit{Junge Freiheit} (\emph{jf}). We also ensure that different outlet types are covered, i.e., daily (\emph{lf, lm, we}) and weekly newspapers (\emph{fo, va, ze}) and a public television channel (\emph{fi, ts}).
Most outlets have an equivalent in Germany/France with regard to e.g., the type of news source or political stance.

\paragraph{Collection Method and Time Frame} 
The \mphasis\ corpus consists of articles and their comment threads. To identify articles on our chosen news articles which are relevant to the topic of migration, we create a list of 8 (\fr) or 9 (\de) migration-related keywords (see Appendix). Note that these keywords are only related to migration and \underline{not} related to HS (i.e., no slurs), and thus leave room for the collection of implicitly hateful comments. We implement a crawler that searches through the outlets' web pages and retrieves an article and its complete comment thread if: $1)$ the article content matches with one of the migration-related keywords; $2)$ was published between January 2020 and May 2020; $3)$ contains at least five comments in its comment thread.
We limit the collected comments in the comment thread of a single article to the 100 most recent comments at the time of crawling. 
Apart from the text, we also collect meta data, i.e., the outlet, title, subtitle, date of publication and author of an article as well as the username and date of a user comment. Each article and comment is assigned a unique ID. Since we reconstruct the hierarchical nature of articles and their comment threads, we also save the ID of the direct parent, i.e., article or other user comment, to which a comment is a direct response. 
The final data sizes per outlet and language are presented in Table \ref{t:outlets}.

\paragraph{Privacy and Copy Right Laws}
To conceal the identity of a user abiding to data privacy laws (GDPR), user names are anonymised using internal user IDs and we do not retain a mapping of the user names to the user IDs. To abide copy right laws, we do not publish the textual content of articles, which are replaced by URLs that point to the corresponding web pages.

\begin{table*}[]
\small
\centering
\begin{tabular}{l|lrr| lrr}
\toprule
\textbf{Type} & \textbf{Outlet (\fr)}         & \textbf{\#Art.} & \textbf{\#Com.} & \textbf{Outlet (\de)}      & \textbf{\#Art.} & \textbf{\#Com.} \\
\midrule
Mainstream           & France Info (\textit{fi})       & 20                & 618             & Tagesschau (\textit{ts})     & 13                & 1,020              \\
                     & Figaro (\textit{lf})            & 19                & 1,056              & Welt (\textit{we})           & 74               & 736              \\
                     & Le Monde (\textit{lm})          & 16                & 554              & Zeit (\textit{ze})           & 15                & 999              \\
                     & Valeurs Actuelles (\textit{va}) & 30                & 614              & Focus (\textit{fo})          & 13                & 962              \\
\midrule
Fringe               & AgoraVox (\textit{av})          & 11                & 369               & Compact (\textit{co})        & 12                & 282               \\
                     & Riposte Laïque (\textit{rl})    & 35                & 1,435              & Epoch Times (\textit{et})    & 2                 & 75               \\
                     &        --                         &    --               &     --              & Junge Freiheit (\textit{jf}) & 55                & 747              \\ \midrule \midrule
Total & -- & 131 & 4,646 & -- & 187 & 4,821 \\ \bottomrule
\end{tabular}
\caption{Overview of number of news articles (\#Art.) and comments (\#Com.) collected from both mainstream and fringe media outlets in French (\fr) and German (\de) for inclusion in the \mphasis\ dataset.}
\label{t:outlets}
\end{table*}

\subsection{Annotation}
\paragraph{Corpus Structure}
To study different types of hate in user comments without an a-priori definition of HS, we develop an annotation protocol that includes various facets of how hate can be communicated in user comments. There are two units of analysis in the corpus: the article and the corresponding comments in its comment thread. 

On the \textbf{article} level, we capture the type of news pieces (\textit{fact} or \textit{opinion oriented}), the topic as well as the first three mentioned main agents (e.g., \textit{politicians, migrants, organisations} etc.). 

For \textbf{comments}, the classes differ between moderation or user comments. For moderation comments (i.e., written by moderators), only the type of moderating action (e.g., \textit{deletion}\footnote{When a comment has been deleted but still shows on the webpage as a comment with the text removed (e.g., \emph{This comment has been deleted.}), this counts as a moderation comment with moderation action \emph{deletion}.}, \textit{referral to the netiquette}, etc.) is annotated. Note that we only have access to publicly available data, thus the proportion of hate comments collected in news outlets with strict moderation policies is affected.
For all user comments, we annotate the topic, potential (\textit{agreeing} or \textit{disagreeing}) references to its parent and the use of amplifiers (i.e., stylistic reinforcing elements). The centrepiece of the corpus is the annotation of HS-related phenomena. Instead of giving annotators a definition of HS, we focus on HS features which in their combination become descriptors of hateful content, described below.

\paragraph{Hate Speech Features}
\begin{figure*}
    \centering
    \includegraphics[width=\linewidth]{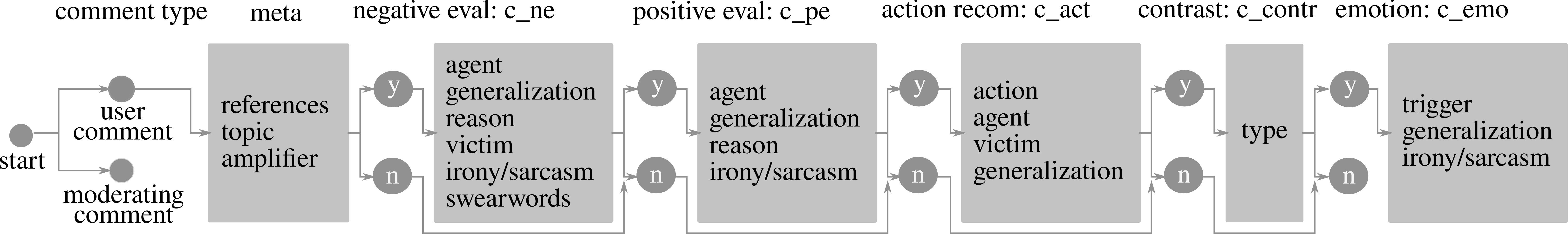}
    \caption{Annotation pipeline for HS features in user comments across five modules (\textttm{c\_ne}, \textttm{c\_pe}, \textttm{c\_act}, \textttm{c\_contr}, \textttm{c\_emo}). When a comment fullfills the requirement (\textit{y}) of a module (e.g., contains negative evaluation for \textttm{c\_ne}), follow-up categories (gray boxes) are annotated, otherwise (\textit{n}) we skip to the next module.}
    \label{f:annotation_pipeline}
\end{figure*}

HS features are annotated across five modules (Figure \ref{f:annotation_pipeline}), each containing 1--7 categories (i.e., \emph{questions}), which can each have several classes (i.e., \emph{answers}). We present these together with an example instance: \textit{Keine Migranten mehr aufnehmen. Wir haben gen\"{u}gend eigene Sorgen.} \translation{No more migrants. We have enough worries of our own.}:

\begin{itemize}
    \item \textttm{c\_ne}: \textbf{Negative evaluation} of an agent, e.g., \textit{migrant, politician}; generalisation-level of agent, i.e., \textit{individual} vs. \textit{group}; whether the evaluation is \textit{explicit} or \textit{implicit}; reason for the evaluation, e.g., \textit{hypocrisy, ignorance, financial burden}; the \textit{victim(s)} of the behaviour of the agent; use of \textit{irony} or \textit{swearwords}. Here in the example: \emph{no negative evaluation}.
    \item \textttm{c\_pe}: \textbf{Positive evaluation} of an agent. The categories are analogous to \textttm{c\_ne}. Here: \emph{none.}
    \item \textttm{c\_act}: \textbf{Recommendation of an action} or behaviour, e.g., \textit{adaption, elimination}; \textit{explicitness} of the recommendation; the \textit{agent} suggested to perform the action and its level of \textit{generalisation}; the \textit{victim(s)} of the action and their level of \textit{generalisation}. Here: \emph{a recommendation to treat migrants (victims), as a group, negatively but violence-free. The agent is unclear.}.
    \item \textttm{c\_contr}: \textbf{Contrasting} between an in- and out-group, e.g., \textit{elite vs. the people}. Here: \emph{migrants vs. German population}.
    \item \textttm{c\_emo}: Expression of a \textit{negative} or \textit{positive} \textbf{emotion}, the trigger of the emotion, e.g., \textit{migrants, media} and its level of \textit{generalisation}; whether the emotion is expressed via \textit{sarcasm}. Here: \emph{no expression of emotion}.
\end{itemize}

Multiple annotations of a single module are possible e.g., if an annotator wants to annotate several negative evaluations.
The HS features above can be combined to create use-case-specific definitions of hate/negativity. For example, the \textttm{c\_ne} module can be used to focus on explicit negative evaluations of groups to describe \emph{explicit HS}, implicit negative evaluations with a reason for \emph{critical comments} or explicit negative evaluations of individuals to focus on \emph{cyber bullying}. Positive evaluations (\textttm{c\_pe}) of controversial groups can also be signs of \emph{HS} or \emph{radicalisation}, and the recommendation of actions such as killing individuals or groups (\textttm{c\_act}) is relevant for identifying HS content illegal in some countries (e.g., according to NetzDG in Germany).

We give more sample annotations and an overview of all annotated modules with their corresponding categories and classes in the Appendix.

\paragraph{Annotators and Annotations}
We recruited four annotators per country. They are native speakers of French/German 
interested in studying HS. They were paid the standard monthly salary of research assistants in Germany/France. 
The annotators went through an extensive training period.
Each instance is annotated by a single annotator and the annotations were performed using HUMAN \cite{wolf2020human}.

At the end of the annotation process, we selected 100 user comments in French and German respectively, which were then annotated by two annotators each to calculate \textbf{inter-annotator agreement} using Brennan and Prediger's Kappa ($\kappa$) \cite{brennan1981coefficient}.
$\kappa$ is calculated for each category, where each individual class per category is treated as a binary \emph{yes}/\emph{no} decision. This makes it possible to calculate the agreement when classes are not mutually exclusive.
Over all categories, we observe high levels of agreement, with all categories being within a reasonable range of $0.77 \leq \kappa \leq 1$. 
We report the inter-annotator agreement values for all categories in the Appendix.

\section{Task-Specific Classification Baselines}
\label{s:baselines}

\begin{table*}[t]
\small
\centering
\begin{tabular}{l l | r r r r r |r r r r r}
\toprule
\textbf{LA} & \textbf{SP} & \textbf{E-1} & \textbf{E-2} & \textbf{E-3a} & \textbf{E-3b} & \textbf{E-3c} & \textbf{A-1} & \textbf{A-2} & \textbf{A-3a} & \textbf{A-3b} & \textbf{A-3c}   \\ \midrule
\de & Train & 2,806 & 1,931 & 1,931 & 1,794 & 1,078 & 2,806 & 624 & 624 & 624 & 624 \\
\de & Dev & 500 & 351 & 351 & 320 & 184 & 500 & 114 & 114 & 114 & 114 \\
\de & Test & 1,000 & 681 & 681 & 632 & 365 & 1,000 & 225 & 225 & 225 & 225 \\ \midrule
\de & $D_{KL}$ & 0.36$_3$ & 0.07$_2$ & 0.20$_5$ & 0.25$_9$ & 0.36$_6$ & 0.16$_2$ & 0.21$_2$ & 0.32$_6$ & 0.56$_5$ & 0.51$_6$ \\ \midrule \midrule
\fr & Train & 2,178 & 1,741 & 1,741 & 1,584 & 1,323 & 2,178 & 680 & 680 & 680 & 680 \\
\fr & Dev & 500 & 409 & 409 & 382 & 206 & 500 & 327 & 327 & 327 & 327 \\
\fr & Test & 1,000 & 795 & 795 & 719 & 607 & 1,000 & 327 & 327 & 327 & 327 \\ \midrule
\fr & $D_{KL}$ & 0.48$_3$ & 0.07$_2$ & 0.24$_5$ & 0.19$_9$ & 0.36$_6$ & 0.07$_2$ & 0.04$_2$ & 0.11$_6$ & 0.33$_5$ & 0.09$_6$ \\
\bottomrule
\end{tabular}
\caption{Number of instances within each sub-task (E, A) in the train, dev and test splits (SP) of the German (\de) and French (\fr) language (LA) corpora. The class imbalance per sub-task is given via the Kullback-Leibler divergence ($D_{KL}$) between the sub-task class distribution of $_c$ classes and a perfectly balanced class distribution.
}
\label{t:task_data}
\end{table*}

\subsection{Experimental Setup}

To provide first insights into the \mphasis\ dataset, we train several baseline models on a number of classification tasks that are based on a subset of classes and categories of the \mphasis\ dataset and analyse their performance and limitations. We focus on two classification tasks, namely task E (i.e., \emph{Evaluation} of agents; based on module \textttm{c\_ne}) and task A (i.e., \emph{Action Recommendation}; module \textttm{c\_act}), which are sentiment-related tasks. 
Each task is divided into 5 sub-tasks to replicate the structure of the \mphasis\ corpus that is based on gradually more in-depth follow-up questions per module. Task \{E$|$A\} have similar structures and are divided into \{E$|$A\}-1 (\emph{Does the comment contain a \{negative or positive evaluation$|$action recommendation\}?}), \{E$|$A\}-2 (\emph{Is the \{evaluation$|$action recommendation\} implicit or explicit?}), \{E$|$A\}-3a (\emph{Who is the target of the \{evaluation$|$action recommendation\}?}), \{E$|$A\}-3b (\emph{What is the \{behaviour of $|$ action recommended to\} the target?}), \{E$|$A\}-3c (\emph{Who is the \{$|$suggested\} victim of the behaviour?}).
We provide a single train(ing), dev(elopment) and test(ing) split.

\paragraph{Data} 
Taking into account the sparsity of most categories in the original \mphasis\ dataset, we create our \textbf{task-based dataset} using only those modules which contain sufficient labelled data. This resulted in using the \textttm{c\_\{ne|pe\}} module for task E and \textttm{c\_act} for task A.
To avoid strong class imbalances within each sub-task, we clustered classes from the original set of classes together which were similar in their underlying meaning and which were sparse in their respective number of instances. We give a more detailed listing of the mapping between the original classes to the sub-tasks' classes in the Appendix. We remove URLs from the text and replace them with a special token, i.e., \textttm{[URL]}. We randomly sample 1,000 and 500 instances from the corpus as the test and dev splits respectively. The remaining 2,806 (\de) or 2,178 (\fr) instances are used for the training data. Sub-tasks \{E$|$A\}-3\{a$|$b$|$c\} are lower-resourced than sub-tasks \{E$|$A\}-\{1$|$2\}, due to their higher number of classes and the smaller amount of available annotations.
For each sub-task, the number of instances of the \de/\fr\ train-dev-test splits and the number of classes are given in Table~\ref{t:task_data}. To give insight into the class imbalance per sub-task, we also report the Kullback-Leibler divergence ($D_{KL}$) between the class distribution of a sub-task and a perfectly balanced class distribution. A rather balanced class distribution would thus lead to a $D_{KL}$ close to $0$.

\paragraph{Model Specifications and Evaluation}
Our baseline models (B) are transformer-based classifiers as implemented in the \textttm{transformers} library.\footnote{\url{https://github.com/huggingface/transformers}} Specifically, we use \texttt{bert-base-german-cased} (\de) and \texttt{camembert-base} \cite{martin-etal-2020-camembert} (\fr). To explore whether domain knowledge can be inserted into the models via intermediate masked-language model (MLM) training, we also fine-tune both language models on their respective \de\ or \fr\ task-based training data for 20 epochs using the MLM objective to obtain task-tuned language models (B+T).
We also explore whether the annotations in the German and French data are sufficiently consistent amongst each other to enable a bilingual learning that improves the classification performance in comparison to a monolingual model.
Therefore, analogous to B+T, we fine-tune a multilingual model \texttt{bert-base-multilingual-cased} on the concatenation of the German and French training data using the MLM objective (M+T) and then learn classification jointly (M+T(J)) or separately (M+T(S)) on the German and French sub-tasks.
All classification models are run over 10 seeded runs with early stopping ($\delta=0.01$, ${\rm patience} = 5$) and we report their average Macro F1 on the test set together with standard mean error. 
For the domain analysis we use the multilingual universal sentence encoder
 \cite{yang-etal-2020-multilingual} to embed user comments, as it works well on semantic similarity tasks \cite{cer2018universal}.

\subsection{Results}

\begin{table*}[t]
\small
\centering
\begin{tabular}{l@{\hskip5pt} l@{\hskip5pt} | r@{\hskip5pt} r@{\hskip5pt} r@{\hskip5pt} r@{\hskip5pt} r@{\hskip5pt} | r@{\hskip5pt} r@{\hskip5pt} r@{\hskip5pt} r@{\hskip5pt} r }
\toprule
\textbf{LA} & \textbf{CM} & \textbf{E-1} & \textbf{E-2} & \textbf{E-3a} & \textbf{E-3b} & \textbf{E-3c} & \textbf{A-1} & \textbf{A-2} & \textbf{A-3a} & \textbf{A-3b} & \textbf{A-3c}   \\ \midrule
\de & B & 55.6$\pm$.5 & 58.7$\pm$.4 & 49.2$\pm$.4 & 27.8$\pm$.9 & 35.2$\pm$.4 & \textbf{72.3$\pm$.4} & 56.2$\pm$1 & \textbf{31.0$\pm$.8}  & 30.8$\pm$.5 & 33.1$\pm$.4 \\
\de & B+T & 55.0$\pm$.2 & 58.6$\pm$.4 & \textbf{51.6$\pm$.6} & \textbf{29.9$\pm$.8} & 35.4$\pm$.3 & 71.3$\pm$.4 & 57.8$\pm$.9 & 28.9$\pm$.9 & 30.8$\pm$1 & \textbf{36.0$\pm$.5} \\ 
\de & M+T(S) & 48.3$\pm$.5 & 52.4$\pm$2 & 45.9$\pm$.5 & 23.4$\pm$.4 & 32.1$\pm$2 & 65.1$\pm$3 & 52.3$\pm$1 & 28.9$\pm$.9 & 28.7$\pm$2 & 28.2$\pm$.7 \\
\de & M+T(J) & 49.0$\pm$1 & 48.1$\pm$4 & 47.5$\pm$.4 & 23.6$\pm$.4 & 34.9$\pm$.7 & 64.1$\pm$2 & 49.5$\pm$2 & 30.7$\pm$1 & 26.8$\pm$.7 & 28.4$\pm$.8 \\ \midrule
\fr & B & 59.3$\pm$.7  & 63.3$\pm$.4 & 54.1$\pm$.5 & 32.9$\pm$.3 & \textbf{39.0$\pm$.3} & 66.9$\pm$.5 & 53.7$\pm$.8 & 40.4$\pm$.5 & 42.1$\pm$.6 & 40.8$\pm$.6 \\
\fr & B+T & 59.6$\pm$.3 & 63.4$\pm$.3 & 53.4$\pm$.4 & 33.5$\pm$.3 & 37.1$\pm$.6 & 67.6$\pm$.3 & 53.2$\pm$.4 & 41.1$\pm$.5 & \textbf{43.8$\pm$.7} & 40.1$\pm$.7 \\
\fr & M+T(S) & 50.3$\pm$1 & 58.8$\pm$.5 & 44.3$\pm$.6 & 23.1$\pm$3 & 32.7$\pm$.4 & 60.2$\pm$2 & 51.2$\pm$1 & 34.5$\pm$.8 & 32.1$\pm$.7 & 34.2$\pm$.9 \\
\fr & M+T(J) & 49.2$\pm$.8 & 49.0$\pm$3 & 45.3$\pm$.6 & 28.0$\pm$.4 & 33.5$\pm$.4 & 51.4$\pm$4 & 52.3$\pm$2 & 36.9$\pm$.6 & 30.6$\pm$2 & 34.4$\pm$.6 \\

\bottomrule
\end{tabular}
\caption{Average Macro F1 of different classification models CM for language LA on the relevant sub-tasks (E,A) test sets. Standard mean errors given as bounds. Top scores outside of the error bounds of other models in \textbf{bold}.}
\label{t:task_results}
\end{table*}

Performing task-based \textbf{intermediate MLM fine-tuning} (B+T) leads to limited improvements over the monolingual baselines (B), with improvements 
up to 
$+2.9$ (\de, A-3c) on the German data (Table \ref{t:task_results}). All improvements are seen on the target-victim sub-tasks \{E$|$A\}-3\{a$|$b$|$c\}. Task domain knowledge acquired by the intermediate MLM training is thus mostly useful for the lower-resourced sub-tasks.
For French, 
most tasks show no significant difference.

The \textbf{multilingual} baselines (M+T) are by far outperformed by their monolingual (B+T) counterparts.
The training on both the French and German data jointly (M+T(J)) leads to some significant improvements on the more complex E-3\{a$|$b$|$c\} sub-tasks in comparison to the multilingual model which was trained on French or German separately (M+T(S)), indicating that there is a sufficient overlap in the French and German annotations such that the lower-resourced sub-tasks benefit from the joint learning; the gain of additional samples outweighing the loss obtained by a few noisy samples.

Overall, we observe low F1 scores across all tasks. This underlines the difficulty of the tasks, which is mostly due to the small amount of samples and sparseness of minority classes, especially for the more complex subtasks. Methods focusing on low-resource classification 
\cite{hedderich-etal-2021-survey}
should be explored to overcome the sparsity in the corpus. 
We give a more detailed account on the error sources in Section \ref{s:qualitative_error_analysis}.

\section{Qualitative Error Analysis}
\label{s:qualitative_error_analysis}

\begin{table*}[t]
\small
\centering
\begin{tabular}{@{\hskip1pt} l | l | l@{\hskip1pt} }
\toprule
EX & Instance & Type \\ \midrule
1 & \textit{Es gibt die ersten Verdachtsfälle in Äthiopien. [...]} \translation{There are some first suspected cases in Ehtiopia. [...]} & $\emptyset \rightarrow N$ \\ \midrule
2 & \textit{Der Berufswunsch dieses jungen Mannes: Politiker! Mehr ist dazu nicht zu sagen.}  \textcolor{gray}{\small{(The career aspiration}} & $\emptyset \rightarrow N$ \\
& \textcolor{gray}{\small{of this young lad: politician! Nothing more to say about this.)}} \\ \midrule
3 & \textit{Man muss sie registrieren (eindeutig, Fingerabdrücke etc!), und Versorgung/Sozialleistungen} & $N \rightarrow \emptyset$ \\
& \textit{gibt's nur am registrierten Ort. Punkt.}  \textcolor{gray}{\small{(They need to be registered (unambiguously, finger prints etc!),}} \\
& \textcolor{gray}{\small{and aid/social benefits only at the registered location. Done.)}} \\ \midrule
4 & \textit{Chouette 2 de moins.} \translation{Cool 2 less.}  &  $N \rightarrow \emptyset$ \\ 
\bottomrule
\end{tabular}
\caption{Example instances (EX) from the DE and FR task E-1 test set with the error type (\textit{reference \textrightarrow\ predicted}) of the best performing classification models in DE (B) and FR (B+T). Classes:  \textit{none} ($\emptyset$), \textit{negative} (N).
}
\label{t:qualitative}
\end{table*}

To further identify shortcomings of the baseline models and difficulties related to the corpus structure, we perform a qualitative error analysis. We focus on the two best models in DE (B) and FR (B+T) on task E-1, as this task focuses on positive/negative evaluations of agents and is thus not far from the popular sentiment analysis task. To this end, we have sampled 100 instances from the DE and FR test set predictions and annotated specific error types (Table \ref{t:qualitative}). 

On the German side, the most common error stems from comments without an evaluation but which were classified as containing a negative evaluation (i.e., \emph{over-blacklisting}), which was prevalent in 18\% of instances. The most common causes for over-blacklisting are $i)$ naming of countries or places (5\%; EX-1), $ii)$ naming of people (especially politicians; 3\%) or $iii)$ other trigger words (e.g., \emph{Nazi, Politiker} \translation{politician}; 4\%; EX-2). This is due to the \textbf{topical bias} in the \mphasis\ corpus. Its focus is on the topic of migration, which is ensured by selecting news articles based on migration-related keywords. This enables the inclusion of comments containing implicit and explicit forms of hate, as well as positive sentiments. 
However, due to this topical focus, politicians are frequent recipients of negative evaluations (Section \ref{s:tuples}), and thus the classifier mistakenly learned to equate the appearance of political actors with a negative sentiment.
While topical bias is not uncommon in HS corpora \cite{wiegand-etal-2019-detection}, 
it should be taken into account when using this data to train models, especially those going into production.

A negative evaluation being ignored by the classifier (i.e., classified as \emph{no evaluation}) is the second most common error (6\%). Mistakes in the annotations are one reason, e.g., in cases where a negative action recommendation was mistakenly annotated as a negative evaluation (EX-3). 
Denoising or similar techniques can be used to mitigate the effects of \textbf{noise} in the annotations. Another source of error stems from the models, which only allow to attribute a single label to each instance. However, in some cases several actors are annotated in the original \mphasis\ corpus with varying evaluations. A \textbf{multi-label} classifier could be used to model this complexity. Lastly, when the negative evaluation is too \textbf{implicit or dependent on context}, the classifier was not able to detect it (2\%; EX-4). The annotators were always shown the context of a given comment (e.g., the article or comment to which the current instance is referring to), which was ignored by our classifiers. Including this contextual information may improve the classification of implicit evaluations.

On the French side, we observe much less cases of over-blacklisting (2\%), while the prevalence of ignoring negative evaluations is the same as for the German model (6\%). The reduced prevalence of over-blacklisting might be due to the larger proportion of fringe media content in the French corpus (44.5\% vs. 22.8\% in Germany), thus 
reducing the amount of neutral/informative content to be mistakenly black-listed.

\section{Target Analysis}
\label{s:tuples}

\begin{figure*}[t]
    \centering
    \includegraphics[width=0.45\linewidth]{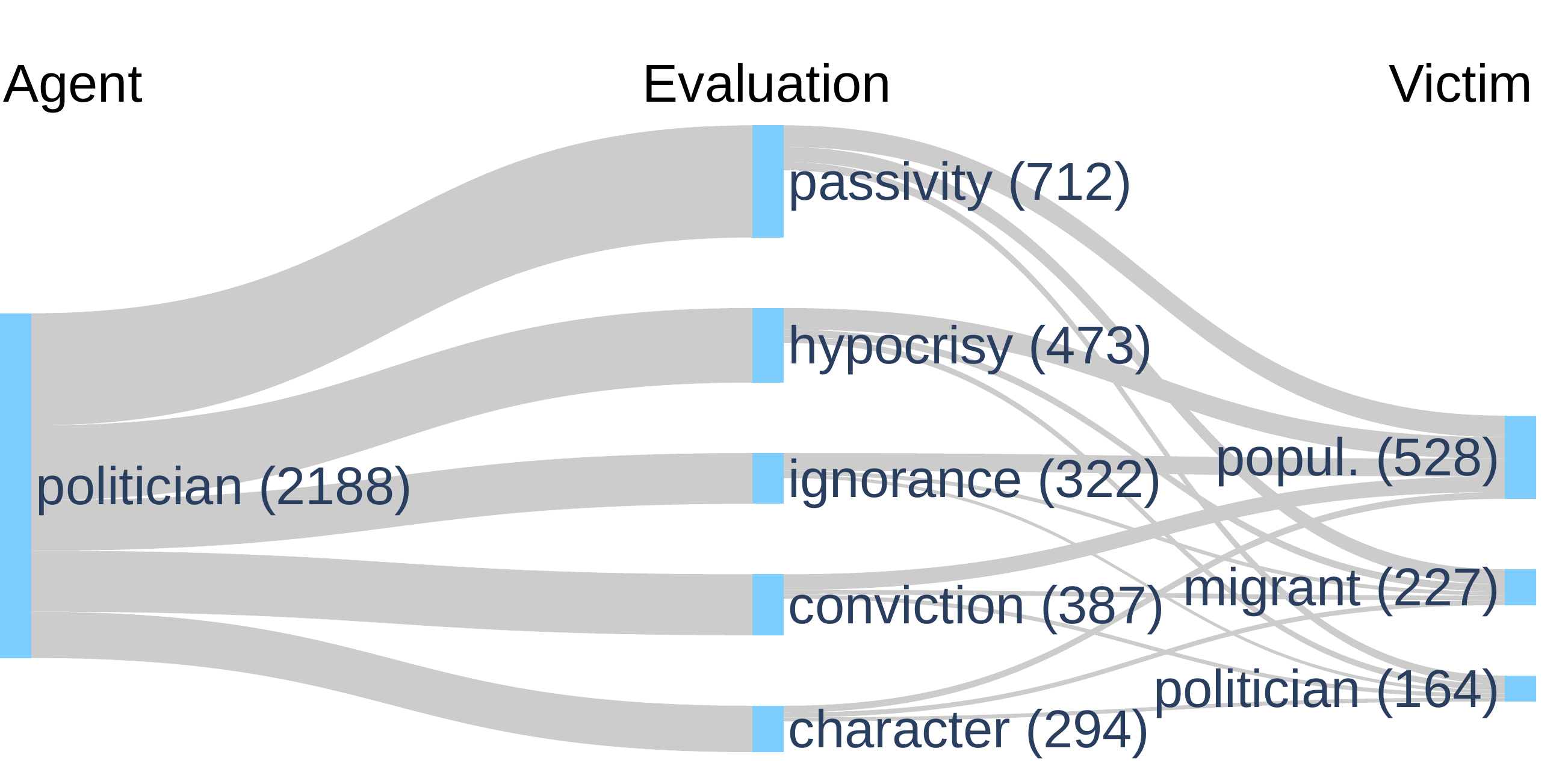}
    \includegraphics[width=0.45\linewidth]{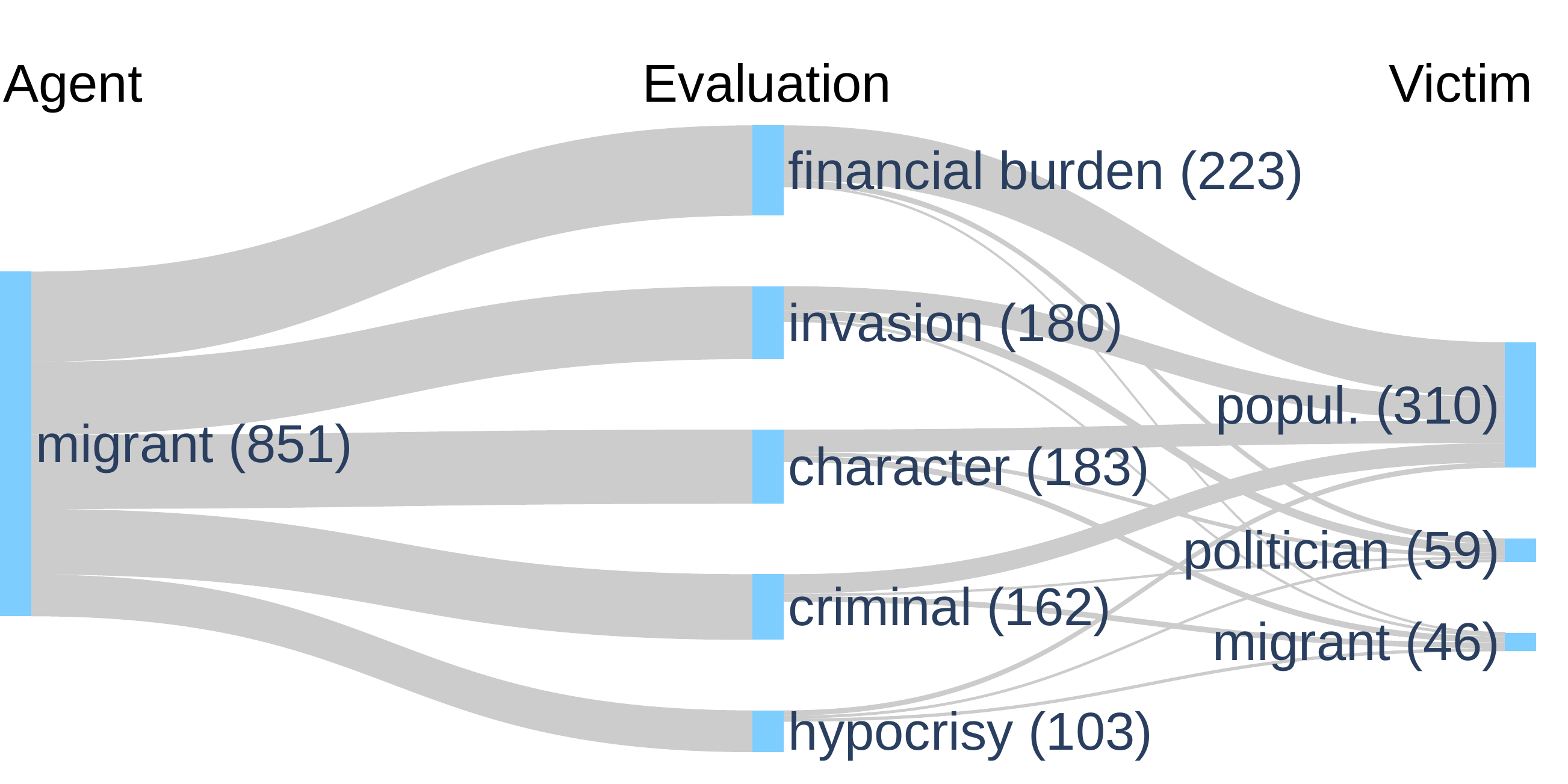}
    \caption{Agent-Evaluation-Victim sankeys for the two most common agents, \emph{politician} (left) and \emph{migrant} (right), in the German portion of the \mphasis\ dataset. We show the 5/3 most common evaluations/victims respectively.}
    \label{fig:agentEvalVictim}
\end{figure*}

\begin{figure*}[t]
    \centering
    \includegraphics[width=0.45\linewidth]{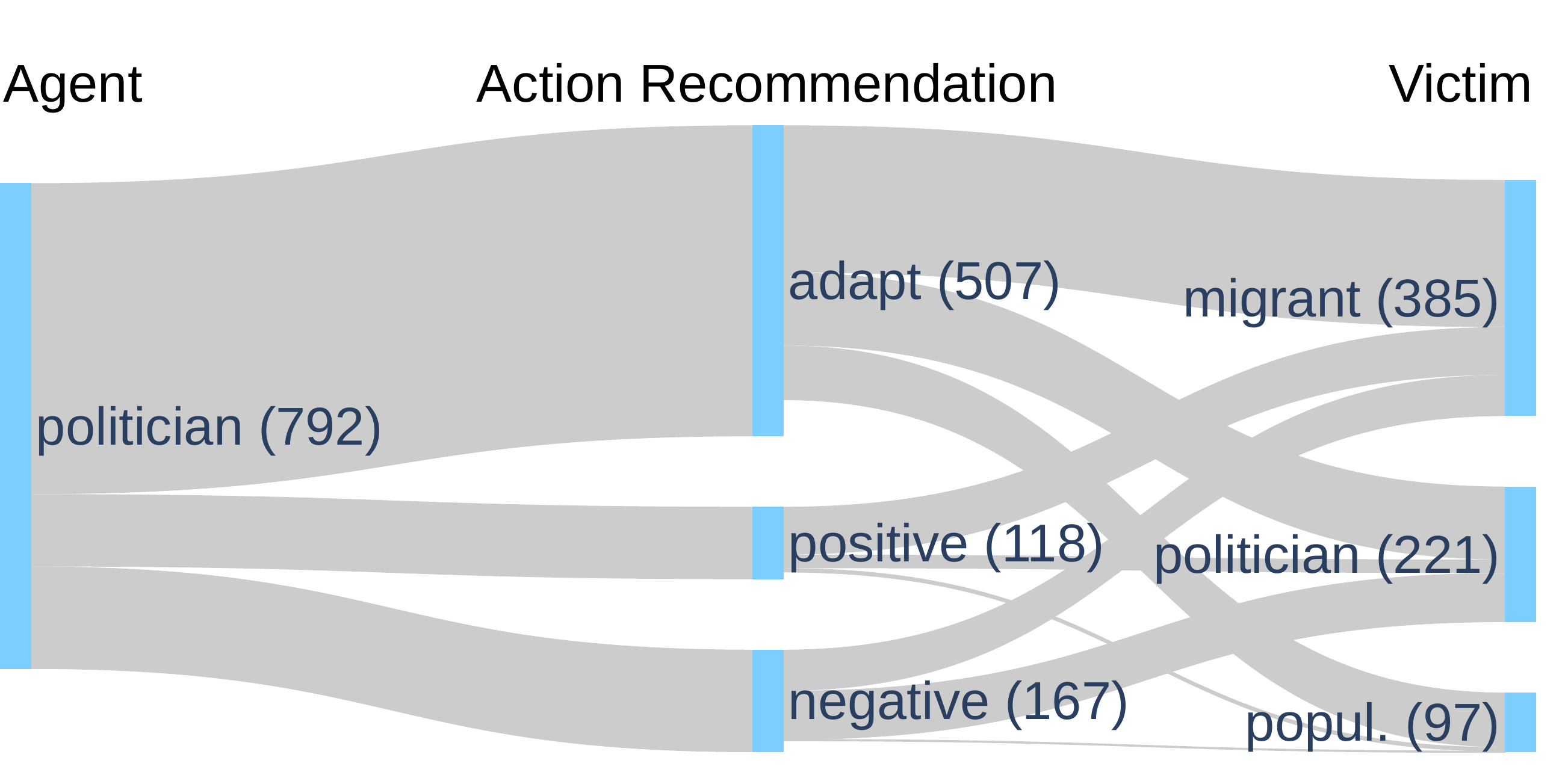}
    \includegraphics[width=0.45\linewidth]{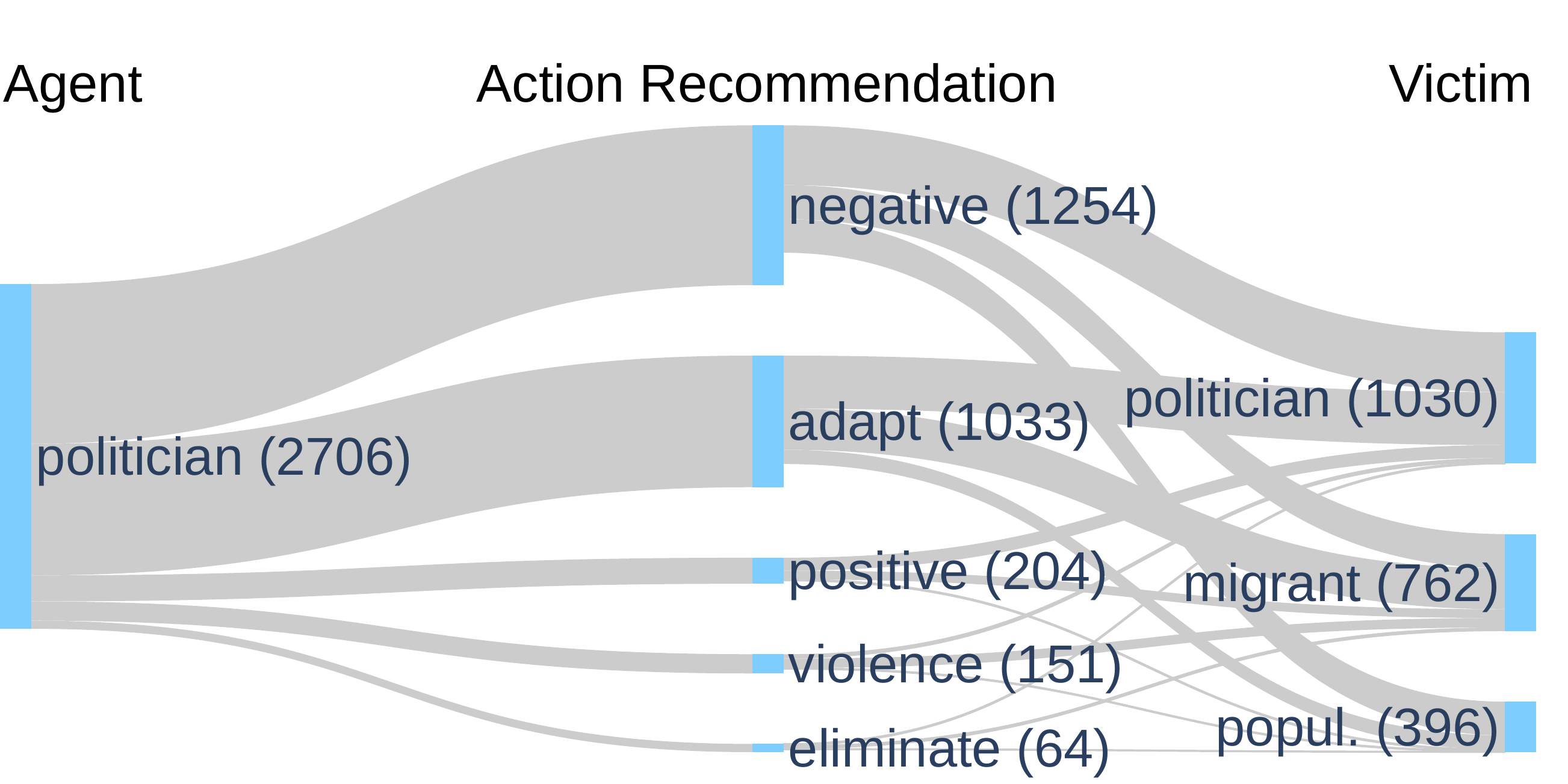}
    \caption{Agent-Action-Victim sankeys for the most common agent (\emph{politician}) in the German (left) and French (right) portions of the \mphasis\ dataset. We show up to 5/3 of the most common evaluations/victims respectively.}
    \label{fig:agentActVictm}
\end{figure*}

One important set of features of the \mphasis\ corpus are the target annotations. This includes the annotation of positive and negative evaluations of targets in user comments (i.e., \textttm{c\_\{ne|pe\}} or task E in Section \ref{s:baselines}). Concretely, a target (\textit{agent}) is evaluated by a user based on their actions (\textit{evaluation}) which have caused harm (or benefit) to a third party (\textit{victim}). Analogous to these agent-evaluation-victim triples, we also obtain agent-action-victim triples (i.e., \textttm{c\_act} or task A), where a user suggests that the agent performs an action (\textit{action}) under which a victim should suffer. 
We explore some of the main trends found in the triple annotations.

For both the German and the French portions of the \mphasis\ corpus, politicians (2.5$k$ (\de) /4.4$k$ (\fr) mentions)\footnote{Note that these numbers are reduced in Figures \ref{fig:agentEvalVictim} and \ref{fig:agentActVictm}, as they only show the most frequent evaluations/actions.} and migrants (1$k$/1.7$k$) were the most common targets of \textbf{negative evaluations}. 
On the German portion (Figure \ref{fig:agentEvalVictim}), the most common mentions of political agents with a negative evaluation were \emph{EU} and \emph{Merkel}.
Indicating a negative sentiment towards the current government and its handling of the topic of migration.
The most frequent negatively evaluated action of these political agents on the German side was \textit{passivity} (712 times).
The two major mentioned victims of the behaviour of politicians are, by a large part (589 times) the German population or, to a smaller extend (299), migrants.
When migrants are the recipients of negative evaluations, the action they are most frequently accused of are being a \emph{financial burden} (223), with the population being the most frequently mentioned victim (353).
Nevertheless, the fact that politicians are by far more frequently negatively evaluated than migrants on the topic of migration, shows that negative sentiments tend to be directed to the decision makers of migration-related policies. This blaming reflects the notion that migrants also suffer from these policies (as a frequent victim group of politicians).

The two most frequent agents to be addressed in an \textbf{action recommendation} by the German users (Figure \ref{fig:agentActVictm} (left)) are either political actors (792) 
or the German population (289).
German politicians are most frequently called to force foreigners to adapt to German society (240) and when they are called to treat someone negatively, the largest victim group are mostly foreign political entities (\textit{EU}, \textit{T\"{u}rkei} \translation{Turkey}, \textit{Griechenland} \translation{Greece}).
When politicians are called to treat someone positively, the suggested beneficiary are most frequently migrants (78). Similar trends are also found in French user comments.
While the German calls for action tend to be more moderate, 
the action recommendations in France (Figure \ref{fig:agentActVictm} (right)) include more radical actions such as physical violence towards (74) or elimination of (29) foreigners. It is unclear whether this more radical manifestation of hate towards migrants is due to an increased societal radicalisation or a difference in the data, where $a)$ German comments are more likely to be published in mainstream media compared to French comments (Section \ref{s:domain_analysis}), or $b)$ the German news outlets are more pro-active in their moderation strategies, deleting more radical comments before we could collect them for the \mphasis\ corpus.

The sparsity of most 
triples makes their prediction using classification models especially difficult, which can be observed in the generally low F1 scores of the baselines on sub-tasks \{E$|$A\}-3\{a$|$b$|$c\} (Table \ref{t:task_results}).

\section{Domain Analysis}
\label{s:domain_analysis}

\begin{figure}[t]
    \centering
    \includegraphics[width=0.82\linewidth]{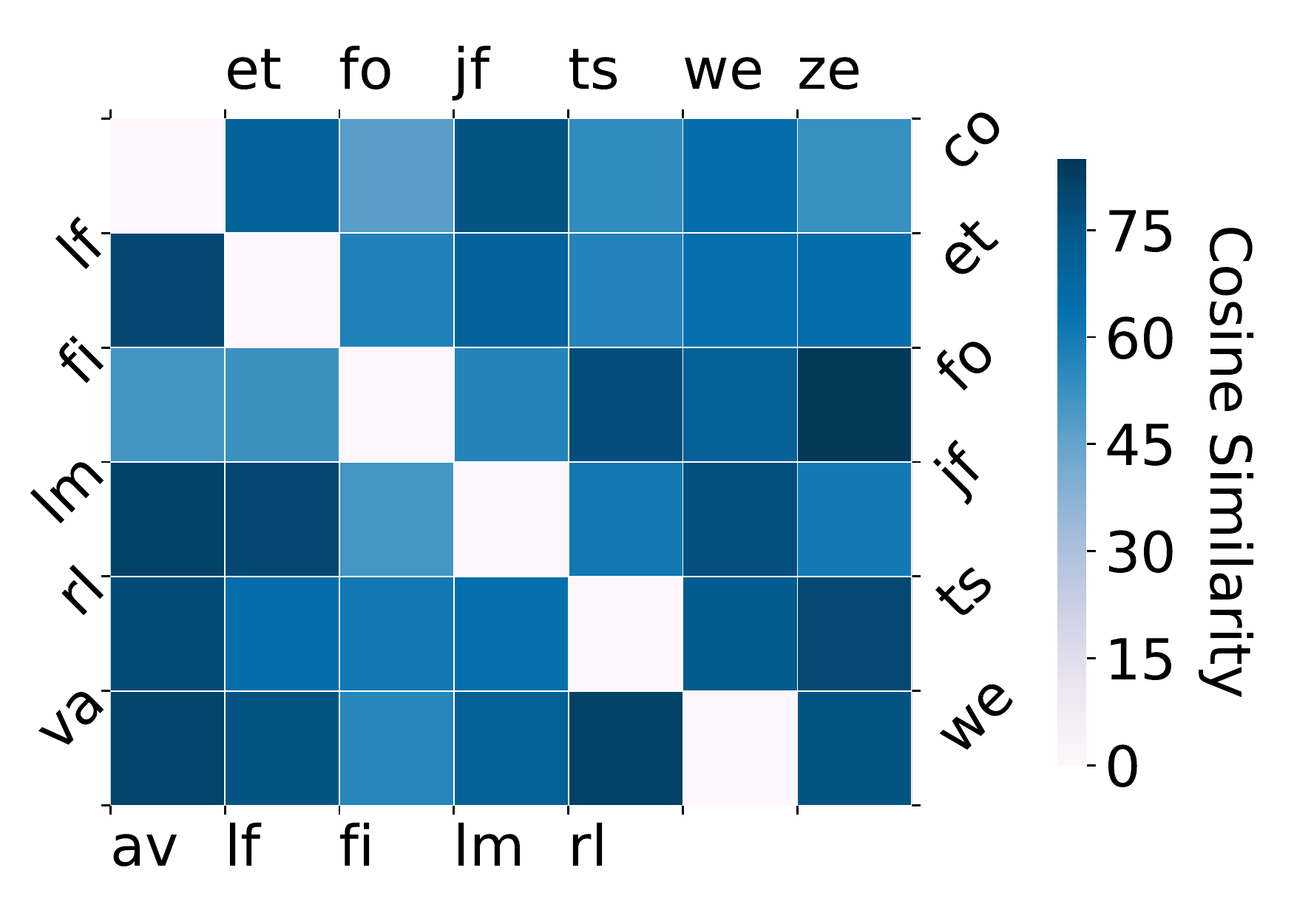}
    \caption{Cosine similarity between universal sentence encoder embeddings of user comments in DE (top-right triangle) or FR (bottom-left triangle) news outlets.}
    \label{f:domain_sims}
\end{figure}

\begin{table}[t]
\small
\centering
\begin{tabular}{l@{\hskip2pt} l@{\hskip3pt} | r@{\hskip5pt} r@{\hskip5pt} r@{\hskip5pt} r@{\hskip5pt} r  }
\toprule
\textbf{TS} & \textbf{CM} & \textbf{E-1} & \textbf{E-2} & \textbf{E-3a} & \textbf{E-3b} & \textbf{E-3c}   \\ \midrule
M & B & 53.8$\pm$.2 & 56.7$\pm$.3 & \textbf{49.7$\pm$.5} & \textbf{26.3$\pm$.8} & \textbf{35.1$\pm$.6} \\
F & B & \textbf{57.1$\pm$.9} & \textbf{61.0$\pm$1} & 46.6$\pm$1 & 24.7$\pm$.1 & 21.6$\pm$2 \\ \midrule
M & B+T & \textbf{53.8$\pm$1} & 56.8$\pm$.2 & 49.8$\pm$.6 & \textbf{26.6$\pm$.7} & \textbf{35.2$\pm$.5} \\
F & B+T & 51.7$\pm$1 & \textbf{59.3$\pm$1} & 50.9$\pm$.7 & 22.4$\pm$.8 & 17.4$\pm$.8 \\
\bottomrule
\end{tabular}
\caption{Average Macro F1 of German classification models CM tested on the mainstream (M) or fringe (F) test set (TS) of the relevant sub-tasks (E). Standard mean errors given as bounds. Top scores outside of the error bounds of other models in \textbf{bold}.
}
\label{t:domain_results}
\end{table}

The \mphasis\ corpus contains user comments from various news outlets. As these outlets differ in political orientation, their user comments also tend to differ in style. This can lead to domain differences across instances, which can affect the performance of classifiers trained on the \mphasis\ dataset. To quantify this domain difference between user comments, we generate embeddings on the concatenation of all user comments belonging to a single news outlet using universal sentence encoder. For each language, we calculate the cosine similarity between the embeddings (Figure \ref{f:domain_sims}).

There is a strong correlation between the manual categorisation of news outlets as either \emph{mainstream} or \emph{fringe} and the similarity between user comments of these two types of media outlets. Specifically, user comments posted in mainstream media such as \textit{Focus}, \textit{Tagesschau}, \textit{Welt} and \textit{Zeit} are closer to each other than to fringe media such as \textit{Compact}, \textit{Epoch Times} or \textit{Junge Freiheit}. Performing K-Means clustering over the German document embeddings ($k=2$) yields exactly the same divide between user comments of mainstream and fringe media. For French, there is a three-step divide ($k=3$), where comments under news outlets are clustered into fringe (\textit{Agoravox}, \textit{Riposte Laïque}, \textit{Valeurs Actuelles}), mainstream (\textit{Le Figaro}, \textit{Le Monde}) and intellectual (\textit{France Info}). 

For the German data $77.2\%$ of comments are from mainstream media, while for the French side the domains are more balanced, with $55.5\%$ of comments stemming from either mainstream or intellectual media. 
To quantify the effect of this domain imbalance on the German data, we evaluate the German B and B+T models on the subset of instances in the test sets that stem from $1)$ mainstream or $2)$ fringe media outlets. We focus this analysis on task E, as it contains more samples across its sub-tasks than task A (Table \ref{t:domain_results}).

For the sparse multi-class sub-tasks E-3\{a$|$b$|$c\}, 
the performance on the more data-rich mainstream comments is comparatively higher (B), underlining the fact that the domain differences in the \mphasis\ corpus are especially to be taken into consideration when working with data sparse classification tasks. With intermediate MLM training (B+T) the macro F1 performance on fringe comments drops across most tasks 
in comparison to the performance without intermediate MLM training (B), indicating that the domain-imbalance on the German data of the \mphasis\ corpus was transmitted into the representations of the underlying BERT model, leading to a lower classification performance on the under-represented fringe domain.

\section{Conclusion and Future Work}

We present \mphasis, a corpus of $\sim9k$ German and French comments collected from migration-related news articles. While most existing HS corpora rely on an ad-hoc definition of HS, which ignores the complex nature of hate online, the \mphasis\ corpus does not attempt to judge whether a user comment is hateful or not. Instead, it focuses on a total of $23$ HS-related features,
which in their combination become descriptors of various types of hateful content. 
We discuss baseline results on several sub-tasks created from the \mphasis\ corpus together with a qualitative error analysis. We analyse evaluations and action recommendations and quantify the domain differences between comments of different sources included in the \mphasis\ corpus.

The \mphasis\ corpus leaves room for various types of analysis. Comments are collected with their context, thus the relation and flow of information between instances can be analysed. The feature-based approach allows for analysis on correlations between different HS features. Comments stem from various news outlets and make a cross-media analysis possible. The bilingual nature of the corpus also allows for a cross-cultural study of HS phenomena in France and Germany.

The \mphasis\ corpus, the train-dev-test splits, model outputs and annotation protocol are made public under \url{https://github.com/uds-lsv/mphasis}.

\section*{Acknowledgments}
We thank our annotators for their keen work.
The project on which this paper is based is funded by the DFG (WI 4204/3-1) and ANR (ANR-18-FRAL-0005).

\section{Bibliographical References}\label{reference}

\bibliographystyle{lrec2022-bib}
\bibliography{anthology,lrec2022-example}

\begin{thebibliography}{}

\bibitem[\protect\citename{Alakrot \bgroup et al.\egroup
  }2018]{azalden2018dataset}
Alakrot, A., Murray, L., and Nikolov, N.~S.
\newblock (2018).
\newblock Dataset construction for the detection of anti-social behaviour in
  online communication in arabic.
\newblock {\em Procedia Computer Science}, 142:174--181.
\newblock Arabic Computational Linguistics.

\bibitem[\protect\citename{Basile \bgroup et al.\egroup
  }2019]{basile-etal-2019-semeval}
Basile, V., Bosco, C., Fersini, E., Nozza, D., Patti, V., Rangel~Pardo, F.~M.,
  Rosso, P., and Sanguinetti, M.
\newblock (2019).
\newblock {S}em{E}val-2019 task 5: Multilingual detection of hate speech
  against immigrants and women in twitter.
\newblock In {\em Proceedings of the 13th International Workshop on Semantic
  Evaluation}, pages 54--63, Minneapolis, Minnesota, USA, June. Association for
  Computational Linguistics.

\bibitem[\protect\citename{Berelson}1952]{berelson1952content}
Berelson, B.
\newblock (1952).
\newblock {\em Content Analysis in Communication Research}.
\newblock Foundations of communication research. Free Press.

\bibitem[\protect\citename{Bose \bgroup et al.\egroup
  }2021]{bose-etal-2021-unsupervised}
Bose, T., Illina, I., and Fohr, D.
\newblock (2021).
\newblock Unsupervised domain adaptation in cross-corpora abusive language
  detection.
\newblock In {\em Proceedings of the Ninth International Workshop on Natural
  Language Processing for Social Media}, pages 113--122, Online, June.
  Association for Computational Linguistics.

\bibitem[\protect\citename{Brennan and Prediger}1981]{brennan1981coefficient}
Brennan, R.~L. and Prediger, D.~J.
\newblock (1981).
\newblock Coefficient kappa: Some uses, misuses, and alternatives.
\newblock {\em Educational and Psychological Measurement}, 41(3):687--699.

\bibitem[\protect\citename{Bretschneider and
  Peters}2017]{Bretschneider2017Detecting}
Bretschneider, U. and Peters, R.
\newblock (2017).
\newblock Detecting offensive statements towards foreigners in social media.
\newblock In {\em HICSS}.

\bibitem[\protect\citename{Cer \bgroup et al.\egroup }2018]{cer2018universal}
Cer, D., Yang, Y., Kong, S., Hua, N., Limtiaco, N., John, R.~S., Constant, N.,
  Guajardo{-}Cespedes, M., Yuan, S., Tar, C., Sung, Y., Strope, B., and
  Kurzweil, R.
\newblock (2018).
\newblock Universal sentence encoder.
\newblock {\em CoRR}, abs/1803.11175.

\bibitem[\protect\citename{Chung \bgroup et al.\egroup
  }2019]{chung-etal-2019-conan}
Chung, Y.-L., Kuzmenko, E., Tekiroglu, S.~S., and Guerini, M.
\newblock (2019).
\newblock {CONAN} - {CO}unter {NA}rratives through nichesourcing: a
  multilingual dataset of responses to fight online hate speech.
\newblock In {\em Proceedings of the 57th Annual Meeting of the Association for
  Computational Linguistics}, pages 2819--2829, Florence, Italy, July.
  Association for Computational Linguistics.

\bibitem[\protect\citename{Coe \bgroup et al.\egroup }2014]{coe2014online}
Coe, K., Kenski, K., and Rains, S.~A.
\newblock (2014).
\newblock Online and uncivil? {P}atterns and determinants of incivility in
  newspaper website comments.
\newblock {\em Journal of Communication}, 64(4):658--679.

\bibitem[\protect\citename{Cohen-Almagor}2018]{Cohen-Almagor_2018}
Cohen-Almagor, R.
\newblock (2018).
\newblock Taking north american white supremacist groups seriously: The scope
  and the challenge of hate speech on the internet.
\newblock {\em International Journal for Crime, Justice and Social Democracy},
  7(2):38--57, Jun.

\bibitem[\protect\citename{D{\"o}ring and Mohseni}2020]{doring2020gendered}
D{\"o}ring, N. and Mohseni, M.~R.
\newblock (2020).
\newblock Gendered hate speech in youtube and younow comments: Results of two
  content analyses.
\newblock {\em SCM Studies in Communication and Media}, 9(1):62--88.

\bibitem[\protect\citename{Erjavec and
  Kova{\v{c}}i{\v{c}}}2012]{erjavec2012you}
Erjavec, K. and Kova{\v{c}}i{\v{c}}, M.~P.
\newblock (2012).
\newblock “you don't understand, this is a new war!” analysis of hate
  speech in news web sites' comments.
\newblock {\em Mass Communication and Society}, 15(6):899--920.

\bibitem[\protect\citename{Harlow}2015]{harlow2015story}
Harlow, S.
\newblock (2015).
\newblock Story-chatterers stirring up hate: Racist discourse in reader
  comments on u.s. newspaper websites.
\newblock {\em Howard Journal Of Communications}, 26:21--42, 01.

\bibitem[\protect\citename{Hedderich \bgroup et al.\egroup
  }2021]{hedderich-etal-2021-survey}
Hedderich, M.~A., Lange, L., Adel, H., Str{\"o}tgen, J., and Klakow, D.
\newblock (2021).
\newblock A survey on recent approaches for natural language processing in
  low-resource scenarios.
\newblock In {\em Proceedings of the 2021 Conference of the North American
  Chapter of the Association for Computational Linguistics: Human Language
  Technologies}, pages 2545--2568, Online, June. Association for Computational
  Linguistics.

\bibitem[\protect\citename{Jha and Mamidi}2017]{jha-mamidi-2017-compliment}
Jha, A. and Mamidi, R.
\newblock (2017).
\newblock When does a compliment become sexist? analysis and classification of
  ambivalent sexism using twitter data.
\newblock In {\em Proceedings of the Second Workshop on {NLP} and Computational
  Social Science}, pages 7--16, Vancouver, Canada, August. Association for
  Computational Linguistics.

\bibitem[\protect\citename{Johnson \bgroup et al.\egroup
  }2019]{johnson2019hidden}
Johnson, N., Leahy, R., Restrepo, N., Velasquez, N., Zheng, M., Manrique, P.,
  Devkota, P., and Wuchty, S.
\newblock (2019).
\newblock Hidden resilience and adaptive dynamics of the global online hate
  ecology.
\newblock {\em Nature}, 573:1--5, 09.

\bibitem[\protect\citename{Jurgens \bgroup et al.\egroup
  }2019]{jurgens-etal-2019-just}
Jurgens, D., Hemphill, L., and Chandrasekharan, E.
\newblock (2019).
\newblock A just and comprehensive strategy for using {NLP} to address online
  abuse.
\newblock In {\em Proceedings of the 57th Annual Meeting of the Association for
  Computational Linguistics}, pages 3658--3666, Florence, Italy, July.
  Association for Computational Linguistics.

\bibitem[\protect\citename{Kovács \bgroup et al.\egroup
  }2021]{kovacs2021challenges}
Kovács, G., Alonso, P., and Saini, R.
\newblock (2021).
\newblock Challenges of hate speech detection in social media.
\newblock {\em SN Computer Science}, 2, 04.

\bibitem[\protect\citename{Mandl \bgroup et al.\egroup
  }2019]{majumder2019overview}
Mandl, T., Modha, S., Majumder, P., Patel, D., Dave, M., Mandlia, C., and
  Patel, A.
\newblock (2019).
\newblock Overview of the hasoc track at fire 2019: Hate speech and offensive
  content identification in indo-european languages.
\newblock In {\em Proceedings of the 11th Forum for Information Retrieval
  Evaluation}, FIRE '19, page 14–17, New York, NY, USA. Association for
  Computing Machinery.

\bibitem[\protect\citename{Martin \bgroup et al.\egroup
  }2020]{martin-etal-2020-camembert}
Martin, L., Muller, B., Ortiz~Su{\'a}rez, P.~J., Dupont, Y., Romary, L., de~la
  Clergerie, {\'E}., Seddah, D., and Sagot, B.
\newblock (2020).
\newblock {C}amem{BERT}: a tasty {F}rench language model.
\newblock In {\em Proceedings of the 58th Annual Meeting of the Association for
  Computational Linguistics}, pages 7203--7219, Online, July. Association for
  Computational Linguistics.

\bibitem[\protect\citename{Ousidhoum \bgroup et al.\egroup
  }2019]{ousidhoum-etal-2019-multilingual}
Ousidhoum, N., Lin, Z., Zhang, H., Song, Y., and Yeung, D.-Y.
\newblock (2019).
\newblock Multilingual and multi-aspect hate speech analysis.
\newblock In {\em Proceedings of the 2019 Conference on Empirical Methods in
  Natural Language Processing and the 9th International Joint Conference on
  Natural Language Processing (EMNLP-IJCNLP)}, pages 4675--4684, Hong Kong,
  China, November. Association for Computational Linguistics.

\bibitem[\protect\citename{Paasch-Colberg \bgroup et al.\egroup
  }2021]{paasch2021insult}
Paasch-Colberg, S., Strippel, C., Trebbe, J., and Emmer, M.
\newblock (2021).
\newblock From insult to hate speech: {M}apping offensive language in {G}erman
  user comments on immigration.
\newblock {\em Media and Communication}, 9(1):171--180.

\bibitem[\protect\citename{Paltoglou \bgroup et al.\egroup
  }2013]{paltoglou2013predicting}
Paltoglou, G., Theunis, M., Kappas, A., and Thelwall, M.
\newblock (2013).
\newblock Predicting emotional responses to long informal text.
\newblock {\em Affective Computing, IEEE Transactions on}, 4:106--115, 01.

\bibitem[\protect\citename{Plaza-Del-Arco \bgroup et al.\egroup
  }2021]{plazadelarco2021multitask}
Plaza-Del-Arco, F.~M., Molina-González, M.~D., Ureña-López, L.~A., and
  Martín-Valdivia, M.~T.
\newblock (2021).
\newblock A multi-task learning approach to hate speech detection leveraging
  sentiment analysis.
\newblock {\em IEEE Access}, 9:112478--112489.

\bibitem[\protect\citename{Robert \bgroup et al.\egroup
  }2016]{faris2016understanding}
Robert, F., Ashar, A., Gasser, U., and Joo, D.
\newblock (2016).
\newblock Understanding harmful speech online.
\newblock {\em Berkman Klein Center for Internet \& Society Research
  Publication}.

\bibitem[\protect\citename{Rossini}2020]{rossini2020beyond}
Rossini, P.
\newblock (2020).
\newblock Beyond incivility: {U}nderstanding patterns of uncivil and intolerant
  discourse in online political talk.
\newblock {\em Communication Research}, page 0093650220921314.

\bibitem[\protect\citename{Rost \bgroup et al.\egroup }2016]{rost2016digital}
Rost, K., Stahel, L., and Frey, B.~S.
\newblock (2016).
\newblock Digital social norm enforcement: Online firestorms in social media.
\newblock {\em PLOS ONE}, 11(6):1--26, 06.

\bibitem[\protect\citename{Ruiter \bgroup et al.\egroup }2019]{ruiter2019lsv}
Ruiter, D., Rahman, M.~A., and Klakow, D.
\newblock (2019).
\newblock Lsv-uds at {HASOC} 2019: The problem of defining hate.
\newblock In Parth Mehta, et~al., editors, {\em Working Notes of {FIRE} 2019 -
  Forum for Information Retrieval Evaluation, Kolkata, India, December 12-15,
  2019}, volume 2517 of {\em {CEUR} Workshop Proceedings}, pages 263--270.
  CEUR-WS.org.

\bibitem[\protect\citename{Saleem \bgroup et al.\egroup }2016]{saleem_web_2016}
Saleem, H.~M., Dillon, K.~P., Benesch, S., and Ruths, D.
\newblock (2016).
\newblock A web of hate: tackling hateful speech in online social spaces.
\newblock In {\em First Workshop on text Analytics for Cybersecurity and Online
  Safety at LREC 2016}.

\bibitem[\protect\citename{Schmidt and
  Wiegand}2017]{schmidt-wiegand-2017-survey}
Schmidt, A. and Wiegand, M.
\newblock (2017).
\newblock A survey on hate speech detection using natural language processing.
\newblock In {\em Proceedings of the Fifth International Workshop on Natural
  Language Processing for Social Media}, pages 1--10, Valencia, Spain, April.
  Association for Computational Linguistics.

\bibitem[\protect\citename{Sigurbergsson and
  Derczynski}2019]{gudbjartur2019offensive}
Sigurbergsson, G.~I. and Derczynski, L.
\newblock (2019).
\newblock Offensive language and hate speech detection for danish.
\newblock {\em CoRR}, abs/1908.04531.

\bibitem[\protect\citename{Silva \bgroup et al.\egroup
  }2021]{silva2021analyzing}
Silva, L., Mondal, M., Correa, D., Benevenuto, F., and Weber, I.
\newblock (2021).
\newblock Analyzing the targets of hate in online social media.
\newblock {\em Proceedings of the International AAAI Conference on Web and
  Social Media}, 10(1):687--690, Aug.

\bibitem[\protect\citename{Sprugnoli \bgroup et al.\egroup
  }2018]{sprugnoli-etal-2018-creating}
Sprugnoli, R., Menini, S., Tonelli, S., Oncini, F., and Piras, E.
\newblock (2018).
\newblock Creating a {W}hats{A}pp dataset to study pre-teen cyberbullying.
\newblock In {\em Proceedings of the 2nd Workshop on Abusive Language Online
  ({ALW}2)}, pages 51--59, Brussels, Belgium, October. Association for
  Computational Linguistics.

\bibitem[\protect\citename{Stru\ss\ \bgroup et al.\egroup
  }2019]{struss2019overview}
Stru\ss\, J.~M., Siegel, M., Ruppenhofer, J., Wiegand, M., and Klenner, M.
\newblock (2019).
\newblock Overview of germeval task 2, 2019 shared task on the identification
  of offensive language.
\newblock Preliminary proceedings of the 15th Conference on Natural Language
  Processing (KONVENS 2019), October 9 – 11, 2019 at
  Friedrich-Alexander-Universit{\"a}t Erlangen-N{\"u}rnberg, pages 352 -- 363,
  M{\"u}nchen [u.a.]. German Society for Computational Linguistics \& Language
  Technology und Friedrich-Alexander-Universit{\"a}t Erlangen-N{\"u}rnberg.

\bibitem[\protect\citename{Su \bgroup et al.\egroup }2018]{su2018uncivil}
Su, L. Y.-F., Xenos, M.~A., Rose, K.~M., Wirz, C., Scheufele, D.~A., and
  Brossard, D.
\newblock (2018).
\newblock Uncivil and personal? {C}omparing patterns of incivility in comments
  on the facebook pages of news outlets.
\newblock {\em New Media \& Society}, 20(10):3678--3699.

\bibitem[\protect\citename{Uyheng and Carley}2021]{uyheng2021characterizing}
Uyheng, J. and Carley, K.
\newblock (2021).
\newblock Characterizing network dynamics of online hate communities around the
  covid-19 pandemic.
\newblock {\em Applied Network Science}, 6, 03.

\bibitem[\protect\citename{Vidgen and Derczynski}2021]{vidgen2021directions}
Vidgen, B. and Derczynski, L.
\newblock (2021).
\newblock Directions in abusive language training data, a systematic review:
  Garbage in, garbage out.
\newblock {\em PLOS ONE}, 15(12):1--32, 12.

\bibitem[\protect\citename{Waseem and Hovy}2016]{waseem2016hateful}
Waseem, Z. and Hovy, D.
\newblock (2016).
\newblock Hateful symbols or hateful people? {P}redictive features for hate
  speech detection on twitter.
\newblock In {\em Proceedings of the NAACL student research workshop}, pages
  88--93.

\bibitem[\protect\citename{Wiegand \bgroup et al.\egroup
  }2019]{wiegand-etal-2019-detection}
Wiegand, M., Ruppenhofer, J., and Kleinbauer, T.
\newblock (2019).
\newblock {D}etection of {A}busive {L}anguage: the {P}roblem of {B}iased
  {D}atasets.
\newblock In {\em Proceedings of the 2019 Conference of the North {A}merican
  Chapter of the Association for Computational Linguistics: Human Language
  Technologies, Volume 1 (Long and Short Papers)}, pages 602--608, Minneapolis,
  Minnesota, June. Association for Computational Linguistics.

\bibitem[\protect\citename{Wolf \bgroup et al.\egroup }2020]{wolf2020human}
Wolf, M., Ruiter, D., D{'}Sa, A.~G., Reiners, L., Alexandersson, J., and
  Klakow, D.
\newblock (2020).
\newblock {HUMAN}: Hierarchical universal modular {AN}notator.
\newblock In {\em Proceedings of the 2020 Conference on Empirical Methods in
  Natural Language Processing: System Demonstrations}, pages 55--61, Online,
  October. Association for Computational Linguistics.

\bibitem[\protect\citename{Wulczyn \bgroup et al.\egroup
  }2017]{wulczyn2017exmachina}
Wulczyn, E., Thain, N., and Dixon, L.
\newblock (2017).
\newblock Ex machina: Personal attacks seen at scale.
\newblock In {\em Proceedings of the 26th International Conference on World
  Wide Web}, WWW '17, page 1391–1399, Republic and Canton of Geneva, CHE.
  International World Wide Web Conferences Steering Committee.

\bibitem[\protect\citename{Yang \bgroup et al.\egroup }2019]{yang2019xlnet}
Yang, Z., Dai, Z., Yang, Y., Carbonell, J., Salakhutdinov, R.~R., and Le, Q.~V.
\newblock (2019).
\newblock Xlnet: Generalized autoregressive pretraining for language
  understanding.
\newblock In {\em Advances in neural information processing systems}, pages
  5753--5763.

\bibitem[\protect\citename{Yang \bgroup et al.\egroup
  }2020]{yang-etal-2020-multilingual}
Yang, Y., Cer, D., Ahmad, A., Guo, M., Law, J., Constant, N., Hernandez~Abrego,
  G., Yuan, S., Tar, C., Sung, Y.-h., Strope, B., and Kurzweil, R.
\newblock (2020).
\newblock Multilingual universal sentence encoder for semantic retrieval.
\newblock In {\em Proceedings of the 58th Annual Meeting of the Association for
  Computational Linguistics: System Demonstrations}, pages 87--94, Online,
  July. Association for Computational Linguistics.

\bibitem[\protect\citename{Ziegele \bgroup et al.\egroup
  }2018]{ziegele2018socially}
Ziegele, M., Koehler, C., and Weber, M.
\newblock (2018).
\newblock Socially destructive? {E}ffects of negative and hateful user comments
  on readers’ donation behavior toward refugees and homeless persons.
\newblock {\em Journal of Broadcasting \& Electronic Media}, 62(4):636--653.

\end{thebibliography}

\bibliographystylelanguageresource{lrec2022-bib}

\section*{Appendix A: List of Keywords}

In order to identify articles related to the topic of migration, we compose a list of regular expression keywords related to this topic. The keywords for France and Germany are equivalent in meaning, however, for Germany there exists one additional keyword, since we include both the more modern and politically correct term \emph{gefl\"{u}chtete*} \translation{refugee} and its older counterpart \emph{fl\"{u}chtling*}, which are both the equivalent of the French keyword \emph{réfugié(s)}.

Concretely, the French keywords are: \textit{étrangers \translation{foreigners}, immigré(s) \translation{immigrant(s)}, migrant(s), réfugié(s), demandeur(s) d’asile \translation{seeker(s) of asylum}, asile, immigration, migration}. 

The German keywords are: \textit{zuwander* \translation{immigrant(s)}, einwander* \translation{immigrant(s)}, migrant*, flüchtling*, geflüchtete*, ausländ* \translation{foreigner(s)}, asyl* \translation{asylum or seeker(s) of asylum}, immigra* \translation{immigration or immigrant(s)}, migration*}.

\section*{Appendix B: Annotation Overview}

We give an overview over the different modules, categories and classes that annotators annotate for each article or comment instance.

\textbf{Articles} are presented to the annotators in the annotation tool without context. Article annotations only have a single module \textttm{article}. The first question (\emph{category}) shown to the annotators is \textttm{n\_2} (Table \ref{t:annotations_articles}), which is always followed by \textttm{n\_3}. If the annotator annotates ${\tt n\_3}=0$ (i.e., \emph{migration not a topic}), then the annotation of the article instance is over and the next instance is shown. If any other class is chosen, then the annotator is asked to also annotate \textttm{n\_4}, where they are asked to choose the first three mentioned agents in the article.

\textbf{Comments} are presented to the annotators together with their direct parent as context, e.g., the news article or another comment to which the current comment is a reply.
As we discern between user comments and moderation comments, the first category shown to annotators is \textttm{c\_usmod} (Table \ref{t:annotations_comments}). If the annotator chooses ${\tt c\_usmod}=0$ (i.e., \emph{moderating comment}), then the annotation of this comment instance is over and the annotation tool proceeds to the next instance. Otherwise, if the comment is annotated as a \emph{user comment}, the annotation tool continues with all follow-up questions in the \textttm{meta} module, i.e., \textttm{c\_refn} to \textttm{c\_amp}. Then, the annotation tool enters the \textttm{c\_ne} module, where negative evaluations are annotated. If the annotator chooses ${\tt c\_ne\_1}=0$, it will skip all following categories in the \textttm{c\_ne} module and jump to the next module \textttm{c\_pe}. Otherwise, it will proceed to ask all dependent follow-up categories \textttm{c\_ne\_2} to \textttm{c\_ne\_7}. At the end of the \textttm{c\_ne} module, the annotation tool asks the annotator whether they want to annotate any further negative evaluations. If this is the case, the tool loops back to the beginning of the \textttm{c\_ne} module. If not, the tool continues to the next module \textttm{c\_pe}, where positive evaluations are annotated. This module functions analogous to the \textttm{c\_ne} module, such that it is only traversed if the annotator states that there is a positive evaluation. Again, multiple traversals are also possible. After the \textttm{c\_pe} module, the annotation tool goes to the \textttm{c\_act} module. If the first category \textttm{c\_act}$=0$, then the tool skips all follow-up categories, otherwise it traverses all categories in the module. Again, several traversals are possible if the annotators choose to annotate several action recommendations. After the \textttm{c\_act} module, there is a single category \textttm{c\_contr}, which also allows multiple answers, followed by the last module, \textttm{c\_emo}. Analogous to previous modules, its dependent categories \textttm{c\_emo\_2a} to \textttm{c\_emo\_3} are only shown if the annotators state that there is an explicit expression of emotions in \textttm{c\_emo\_1}.

\section*{Appendix C: Inter-Annotator Agreement}

\begin{table}[t]
\small
\centering
\begin{tabular}{l|rr|rr}
\toprule
\textbf{Category} & \multicolumn{2}{c}{\textbf{\de}} & \multicolumn{2}{c}{\textbf{\fr}} \\
& \textit{agg} &  $\kappa$ & \textit{agg} & $\kappa$ \\ \midrule
\textttm{c\_usmod} & 1.00 & 1.0 & -- & -- \\
\textttm{c\_refn} & 0.89 & 0.87 & 0.83 & 0.79 \\
\textttm{c\_refc} & 0.96 & 0.94  & 0.93 & 0.90 \\
\textttm{c\_topic} & 0.90 & 0.89 & 0.91 & 0.90 \\
\textttm{c\_amp} & 0.96 & 0.94 & 0.88 & 0.78 \\ \midrule
\textttm{c\_ne\_1} & 0.96 & 0.95 & 0.96 & 0.95 \\
\textttm{c\_ne\_2} & 0.96 & 0.95 & 0.95 & 0.95 \\
\textttm{c\_ne\_3} & 0.99 & 0.99 & 0.99 & 0.99 \\
\textttm{c\_ne\_4} & 0.99 & 0.99 & 0.99 & 0.99 \\
\textttm{c\_ne\_5} & 0.99 & 0.99 & 0.99 & 0.99 \\
\textttm{c\_ne\_6} & 0.99 & 0.99 & 0.99 & 0.99 \\
\textttm{c\_ne\_7} & 0.99 & 0.99 & 0.99 & 0.99 \\ \midrule
\textttm{c\_pe\_1} & 0.84 & 0.80 & 0.83 & 0.77 \\
\textttm{c\_pe\_2} & 0.99 & 0.98 & 0.98 & 0.98 \\
\textttm{c\_pe\_3} & 0.99 & 0.99 & 1.0 & 1.0 \\
\textttm{c\_pe\_4} & 0.99 & 0.99 & 0.99 & 0.99 \\
\textttm{c\_pe\_5} & 1.0 & 1.0 & 1.0 & 1.0 \\ \midrule
\textttm{c\_act} & 0.96 & 0.96 & 0.93 & 0.93 \\
\textttm{c\_act\_1} & 0.95 & 0.98 & 0.92 & 0.92 \\
\textttm{c\_act\_2a} & 0.98 & 0.95 & 0.97 & 0.97 \\
\textttm{c\_act\_2b} & 0.96 & 0.95 & 0.89 & 0.88 \\
\textttm{c\_act\_3a} & 0.98 & 0.98 & 0.97 & 0.97 \\
\textttm{c\_act\_3b} & 0.93 & 0.92 & 0.90 & 0.89 \\ \midrule
\textttm{c\_contr} & 0.96 & 0.96 & 0.94 & 0.93 \\ \midrule
\textttm{c\_emo\_1} & 0.97 & 0.97 & 0.83 & 0.77 \\
\textttm{c\_emo\_2a} & 0.99 & 0.98 & 0.94 & 0.94\\
\textttm{c\_emo\_2b} & 0.98 & 0.97 & 0.82 & 0.76 \\
\textttm{c\_emo\_3} & 0.98 & 0.97 & 0.83 & 0.77\\ \bottomrule
\end{tabular}
\caption{Average agreement (\textit{agg}) and Brennen and Prediger's Kappa ($\kappa$) across all classes in a given category for the German (\de) and French (\fr) side.}
\label{t:agreement}
\end{table}

The 100 user comments selected for calculating the inter-annotator agreement were collected from 5 different articles in German and French respectively. Each user comment is annotated by two annotators. We use Brennen annd Prediger's Kappa ($\kappa$) and the percentage agreement ($agg$) to calculate the inter-annotator agreement. The two metrics are calculated for each category, where each individual class per category is treated as a binary \emph{yes}/\emph{no} decision. Some categories are dependent of other categories, i.e., if an annotator annotates \textttm{c\_ne\_2}$= 1$ (\emph{determinant}), then all following categories  \textttm{c\_ne\_\{3-7\}} (\emph{dependants}) are follow-up questions based on the choice taken in \textttm{c\_ne\_2} (e.g., \emph{is the agent a group or an individual?}). Thus, we only compare the annotations of two annotators on these dependent categories, if they share the same determinant annotation. The determinants are always compared to each other. The determinant \textrightarrow\ dependants groups in the annotations are \textttm{c\_ne\_2} \textrightarrow\ \textttm{c\_ne\_\{1|3-7\}}, \textttm{c\_pe\_2} \textrightarrow\ \textttm{c\_pe\_\{1|3-5\}} and \textttm{c\_act\_1} \textrightarrow\ \textttm{c\_act(\_\{2a|2b|3a|3b\})}.

The average agreement and $\kappa$ per category is reported in Table \ref{t:agreement}.

\section*{Appendix D: Sample Annotations}
We show one example annotation from each country in the corpus to give an intuition how the different annotation modules and categories are applied.

\begin{quote}
    Und Dumm-Michel darf diese Migranten finanzieren. Unglaublich!! \translation{And stupid Michel has to finance these migrants. Incredible!!}
\end{quote}

The above sample instance is a German user comment replying to another comment written under an article talking about new migrants arriving in Berlin. It contains an amplifier (\emph{!!}) (i.e.,\textttm{c\_amp}$ = 1$). It contains an explicit negative evaluation of migrants as a group (\textttm{c\_ne\_1}$ = 1$, \textttm{c\_ne\_2}$ = 1$, \textttm{c\_ne\_3}$ = 2$), and the reason for the evaluation is them being a financial burden (\textttm{c\_ne\_4}$ = 5$). The German population (\emph{Dumm-Michel}) is the victim of this behaviour (\textttm{c\_ne\_5}$ = 1$). There is no sarcasm (\textttm{c\_ne\_6}$ = 0$) or swearwords (\textttm{c\_ne\_7}$ = 0$). There is no positive evaluation (\textttm{c\_pe\_1}$ = 0$), which is why all follow-up categories in the \textttm{c\_pe} module are skipped during annotation. Similarly, there is no action recommendation (\textttm{c\_act}$ = 0$), contrasting (\textttm{c\_contr}$ = 0$) or expression of emotion (\textttm{c\_emo\_1}$ = 0$).

\begin{quote}
    Qu'ils les renvoient en Asie. La frontière est proche. \translation{They should send them back to Asia. The border is close.}
\end{quote}

The above sample instance is a French user comment referring to an article talking about new refugee camps opening on Lesbos and Chios (Greece). It contains no amplifier (\textttm{c\_amp}$ = 0$), no negative evaluation (\textttm{c\_ne\_1}$ = 0$) and no positive evaluation (\textttm{c\_pe\_1}$ = 0$). It does contain an explicit action recommendation (\textttm{c\_act}$ = 1$), namely a negative violence-free treatment (\emph{Qu'ils les renvoient en Asie.}) (\textttm{c\_act\_1}$ = 3$). The suggested agent of the action recommendation is unclear (\textttm{c\_act\_2a}$ = 99$) and the victims are migrants (\textttm{c\_act\_3a}$ = 1$) as a group (\textttm{c\_act\_3b}$ = 2$). There is no contrasting (\textttm{c\_contr}$ = 0$) or expression of emotion (\textttm{c\_emo\_1}$ = 0$).

\section*{Appendix E: Class Mapping}

For the task-specific classification tasks, we select a subset of the \mphasis\ categories and their classes. 
In Table \ref{t:mapping} we list the mapping of \mphasis\ categories and classes to (sub-)tasks and their classes.

\begin{table*}[t]
\tiny
\centering
\begin{tabular}{llll}
\toprule
\textbf{Sub-Task} & \textbf{Class} & \textbf{Class Description} & \textbf{Original Category and Classes} \\ \midrule
E-1 & 0 & negative & \textttm{c\_ne\_1}$ = \{1 | 2\}$ \\
E-1 & 1 & positive & \textttm{c\_pe\_1}$ = \{1 | 2\}$ \\
E-1 & 2 & none & \textttm{c\_ne\_1}$ = 0$ \& \textttm{c\_ne\_0}$ = 0$ \\ \midrule
E-2 & 0 & implicit & \textttm{c\_ne\_1}$ = 2$ \\
E-2 & 1 & explicit & \textttm{c\_ne\_1}$ = 1$ \\ \midrule
E-3a & 0 & migrant & \textttm{c\_ne\_2}$ = \{1 | 111\}$ \\
E-3a & 1 & politician & \textttm{c\_ne\_2}$ = \{2 | 211 - 225\}$ \\
E-3a & 2 & population &  \textttm{c\_ne\_2}$ = 3$ \\
E-3a & 3 & discussants &  \textttm{c\_ne\_2}$ = 13$ \\
E-3a & 4 & other &  \textttm{c\_ne\_2}$ = \{4 - 12\}$ \\ \midrule
E-3b & 0 & passivity & \textttm{c\_ne\_4}$ = 1$ \\
E-3b & 1 & conspiracy & \textttm{c\_ne\_4}$ = 2$ \\
E-3b & 2 & ignorance & \textttm{c\_ne\_4}$ = 3$ \\
E-3b & 3 & criminal behavior & \textttm{c\_ne\_4}$ = 4$ \\
E-3b & 4 & financial burden & \textttm{c\_ne\_4}$ = 5$ \\
E-3b & 5 & incompatibility & \textttm{c\_ne\_4}$ = 6$ \\
E-3b & 6 & invasion & \textttm{c\_ne\_4}$ = 7$ \\
E-3b & 7 & character trait & \textttm{c\_ne\_4}$ = 20$ \\
E-3b & 8 & political conviction & \textttm{c\_ne\_4}$ = 30$ \\ \midrule
E-3c & -- & -- & same mapping as E-3a but with \textttm{c\_ne\_5} \\ \midrule
A-1 & 0 & no & \textttm{c\_act}$ = 0$ \\
A-1 & 1 & yes & \textttm{c\_act}$ = \{1 | 2\}$ \\ \midrule
A-2 & 0 & implicit & \textttm{c\_act}$ = 1$ \\
A-2 & 1 & explicit & \textttm{c\_act}$ = 2$ \\ \midrule
A-3a & -- & -- & same mapping as E-3a but with \textttm{c\_act\_2a} \\
A-3b & 0 & positive treatment & \textttm{c\_act\_1}$ = 1$ \\
A-3b & 1 & adaption & \textttm{c\_act\_1}$ = 2$ \\
A-3b & 2 & negative violence-free & \textttm{c\_act\_1}$ = 3$ \\
A-3b & 3 & physical violence & \textttm{c\_act\_1}$ = 4$ \\
A-3b & 4 & elimination & \textttm{c\_act\_1}$ = 5$ \\ \midrule
A-3c & -- & -- & same mapping as E-3a but with \textttm{c\_act\_3a} \\ \bottomrule
\end{tabular}
\caption{The mapping of sub-task classes to their corresponding original category and class(es).}
\label{t:mapping}
\end{table*}

\begin{table*}[t]
\tiny
\centering
\begin{tabular}{lllllrr}
\toprule
Module & Category & Description & Class Code & Class Description & \#Samples (DE) & \#Samples (FR) \\ \midrule
\textttm{article} & \textttm{n\_2} & type of news piece & 1 & emphasizing facts & 3,626 & 3,030 \\
& & & 2 & emphasizing an opinion & 1,857 & 1,960 \\
& \textttm{n\_3} & topic of news piece & 0 & migration not a topic & 571 & 917 \\
& & & 1 & management of immigration & 2,370 & 1,346 \\
& & & 2 & security and safety & 873 & 888 \\
& & & 3 & justice & 63 & 0 \\
& & & 4 & integration and cohabitation & 22 & 527 \\
& & & 5 & culture and religion & 131 & 74 \\
& & & 6 & education & 0 & 0 \\
& & & 7 & labor market and economy & 530 & 105 \\
& & & 8 & social issues & 371 & 113 \\
& & & 9 & health aspects & 93 & 178 \\
& & & 10 & environment & 14 & 0 \\
& & & 11 & media coverage on migration & 246 & 0 \\
& & & 99 & cannot tell & 219 & 842 \\
& \textttm{n\_4} & mentioned agents & 1 & migrants & 4,538 & 4,724 \\
& & & 111 & residents of other countries & 369 & 400 \\
& & & 2 & in the area of politics & 250 & 307 \\
& & & 211 & CDU/CSU - LR & 1,135 & 17 \\
& & & 212 & SPD - PS & 591 & 8 \\
& & & 213 & B\"{u}ndnis 90/Die Gr\"{u}nen - Les Verts & 396 & 0 \\
& & & 214 & Left-wing politicians & 107 & 61 \\
& & & 215 & FDP - En Marche! & 0 & 0 \\
& & & 216 & AfD - RN/FN & 833 & 773 \\
& & & 217 & government & 833 & 773 \\
& & & 218 & opposition & 0 & 0 \\
& & & 219 & left-wing political camp & 11 & 0 \\
& & & 220 & right-wing political camp & 52 & 0 \\
& & & 221 & left-wing extremists & 45 & 25 \\
& & & 222 & right-wing extremists & 911 & 0 \\
& & & 223 & political and public institutions & 5,478 & 1,669 \\
& & & 224 & states & 3,325 & 670 \\
& & & 225 & foreign politician/party/government & 1,527 & 2,188 \\
& & & 3 & the German/French population & 580 & 403 \\
& & & 4 & media & 406 & 865 \\
& & & 5 & civil society actor & 1,294 & 3,526 \\
& & & 6 & religious actors & 452 & 232 \\
& & & 7 & scientific actors & 68 & 32 \\
& & & 8 & police & 541 & 798 \\
& & & 9 & courts & 568 & 343 \\
& & & 10 & military & 244 & 0 \\
& & & 12 & abstract entities (values, practices etc.) & 0 & 0 \\
& & & 13 & discussant & 0 & 0 \\
& & & 99 & cannot tell & 193 & 86 \\ \bottomrule
\end{tabular}
\caption{Annotation modules for articles and their respective categories. Labels for each category are given with the corresponding number of German (DE) and French (FR) user comments that are part of the comment thread of an article with the given label.}
\label{t:annotations_articles}
\end{table*}

\begin{table*}[t]
\tiny
\centering
\begin{tabular}{lllllrr}
\toprule
Module & Category & Description & Label Code & Label Description & \#Samples (DE) & \#Samples (FR) \\ \midrule
\textttm{meta} & \textttm{c\_usmod} & type of comment & 0 & moderating comment & 76 & 27 \\
& & & 1 & user comment & 4,745 & 3,910 \\
& \textttm{c\_refn} & reference to news article & 0 & makes no reference to news article & 874 & 1,079 \\
& & & 1 & approval of the article & 132 & 118 \\
& & & 2 & refusal of the article & 438 & 252 \\
& & & 3 & ambivalent & 121 & 42 \\
& & & 4 & establishes reference, without evaluating it & 3,102 & 2,377 \\
& & & 99 & cannot tell & 79 & 49 \\
& \textttm{c\_refc} & reference to comment & 0 & does not refer to another comment & 2,757 & 2,375 \\
& & & 1 & agreement & 829 & 863 \\
& & & 2 & disagreement & 1,160 & 677 \\
& & & 99 & cannot tell & 0 & 1,075 \\
& \textttm{c\_topic} & topic of comment & same as \textttm{n\_3} \\
& \textttm{c\_amp} & amplifier & 0 & no & 4,230 & 3,338 \\
& & & 1 & yes & 512 & 576 \\
& & & 99 & cannot tell & 4 & 1 \\ \midrule
\textttm{c\_ne} (negative evaluations) & \textttm{c\_ne\_1} & negative evaluation & 0 & no & 1,351 & 744 \\
& & & 1 & yes, explicit & 2,148 & 2,093 \\
& & & 2 & yes, implicit & 1,083 & 1,024 \\
& & & 99 & cannot tell & 164 & 1,077 \\
& \textttm{c\_ne\_2} & agent & & same as \textttm{n\_4} \\
& \textttm{c\_ne\_3} & level of generalization & 1 & case-specific & 997 & 1,328 \\
& & & 2 & generalized entity & 2,218 & 1,808 \\
& & & 99 & cannot tell & 180 & 33 \\
& \textttm{c\_ne\_4} & reason for the evaluation & 1 & passivity & 604 & 604 \\
& & & 2 & conspiracy or hypocrisy & 778 & 604 \\
& & & 3 & ignorance & 378 & 301 \\
& & & 4 & criminal behaviour & 243 & 309 \\
& & & 5 & financial burden & 148 6 143 \\
& & & 6 & incompatibility & 23 & 117 \\
& & & 7 & invasion & 75 & 185 \\
& & & 8 & illness & 12 & 39 \\
& & & 20 & character traits & 350 & 289 \\
& & & 30 & political conviction & 440 & 151 \\
& & & 99 & cannot tell & 344 & 271 \\
& \textttm{c\_ne\_5} & victim of behavior & & same as \textttm{n\_4} \\
& \textttm{c\_ne\_6} & irony or sarcasm & & same as \textttm{c\_amp} \\
& \textttm{c\_ne\_7} & swearwords & & same as \textttm{c\_amp} \\ \midrule
\textttm{c\_pe} (positive evaluation) & \textttm{c\_pe\_1} & positive evaluation & & same as \textttm{c\_ne\_1} \\
& \textttm{c\_pe\_2} & agent & & same as \textttm{c\_ne\_2} \\
& \textttm{c\_pe\_3} & level of generalization & & same as \textttm{c\_ne\_3} \\
& \textttm{c\_pe\_4} & reason for the evaluation & 1 & efficiency & 118 & 248 \\
& & & 2 & honesty & 30 & 32 \\
& & & 3 & seeing things through & 77 & 111 \\
& & & 4 & exemplary behavior & 69 & 83 \\
& & & 5 & financial advantages & 17 & 19 \\
& & & 6 & cultural enrichment & 2 & 52 \\
& & & 20 & character traits & 30 & 64 \\
& & & 30 & political conviction & 62 & 40 \\
& & & 99 & cannot tell & 28 & 44 \\
& \textttm{c\_pe\_5} & irony or sarcasm & & same as \textttm{c\_ne\_6} \\ \midrule
\textttm{c\_act} (action) & \textttm{c\_act} & action recommendation & 0 & no action & 3,708 & 2,647 \\
& & & 1 & explicit action & 834 & 817 \\
& & & 2 & implicit action & 203 & 451 \\
& \textttm{c\_act\_1} & action & 1 & positive treatment & 172 & 100 \\
& & & 2 & call for change/adaption & 591 & 446 \\
& & & 3 & negative but violence free treatment & 229 & 552 \\
& & & 4 & physical violence & 15 & 60 \\
& & & 5 & elimination/killing & 8 & 90 \\
& & & 99 & cannot tell & 22 & 20 \\
& \textttm{c\_act\_2a} & agent & & same as \textttm{n\_4} \\
& \textttm{c\_act\_2b} & level of generalization & & \textttm{c\_ne\_3} \\
& \textttm{c\_act\_3a} & victim & & same as \textttm{n\_4} \\
& \textttm{c\_act\_3b} & level of generalization & & \textttm{c\_ne\_3} \\ \midrule
\textttm{c\_contr} (contrasting) & \textttm{c\_contr} & contrasted groups & 0 & none & 3,760 & 2,030 \\
& & & 1 & elite vs. the people & 191 & 638 \\
& & & 2 & globalism vs. states & 5 & 41 \\
& & & 3 & right-wing vs. left-wing camps & 40 & 58 \\
& & & 4 & less advantaged citizens vs. migrants & 24 & 28 \\
& & & 5 & french vs. migrants & 1 & 483 \\
& & & 6 & germans vs. migrants & 134 & 4 \\
& & & 7 & french vs. other political actors abroad & 1 & 88 \\
& & & 8 & germans vs. other political actors abroad & 134 & 0 \\
& & & 9 & europeans/westerners vs. others & 75 & 186 \\
& & & 10 & pro-migrants vs. anti-migrants & 46 & 95 \\
& & & 11 & good migrants vs. bad migrants & 55 & 10 \\
& & & 12 & present vs. past & 76 & 28 \\
& & & 99 & cannot tell & 200 & 222 \\ \midrule
\textttm{c\_emo} (emotion) & \textttm{c\_emo\_1} & explicit expression of emotion & 0 & none & 4,577 & 3,429 \\
& & & 1 & negative emotion & 110 & 365 \\
& & & 2 & positive emotion & 38 & 105 \\
& & & 3 & expression of amusement/ridiculing & 21 & 0 \\
& & & 99 & cannot tell & 21 & 16 \\
& \textttm{c\_emo\_2a} & trigger for emotion & & same as \textttm{n\_4} \\
& \textttm{c\_emo\_2b} & level of generalization & & same as \textttm{c\_ne\_3} \\
& \textttm{c\_emo\_3} & irony or sarcasm & & same as \textttm{c\_ne\_6} \\ \bottomrule
\end{tabular}
\caption{Annotation modules for comments and their respective categories. Labels for each category with their corresponding number of German (DE) and French (FR) comments.}
\label{t:annotations_comments}
\end{table*}

\end{document}